\documentclass[lettersize,journal]{IEEEtran}
\usepackage{amsmath,amsfonts}
\usepackage{algorithmic}
\usepackage{array}
\usepackage[caption=false,font=footnotesize]{subfig}
\usepackage{booktabs}
\usepackage{textcomp}
\usepackage{stfloats}
\usepackage{url}
\usepackage{verbatim}
\usepackage{graphicx}
\usepackage{adjustbox}
\usepackage{multirow}
\usepackage{xcolor}

\def\BibTeX{{\rm B\kern-.05em{\sc i\kern-.025em b}\kern-.08em
    T\kern-.1667em\lower.7ex\hbox{E}\kern-.125emX}}
\usepackage{balance}

\usepackage[colorlinks=true, citecolor=blue, linkcolor=blue, urlcolor=blue]{hyperref}
\usepackage{orcidlink}
\begin{document}
\title{HOSE: Hierarchical Orthogonal Subspace Erasure for Concept Unlearning in Diffusion Models}
\author{Xiaoqi Wei, 
Li Li\,\orcidlink{0000-0003-4799-5896}, 
Kejiang Chen\,\orcidlink{0000-0002-9868-3414}, 
Hanzhou Wu\,\orcidlink{0000-0002-1599-7232}, 
and Xinpeng Zhang%
\thanks{This work was supported by the National Natural Science Foundation of China under Grant 62302286, and the Xizang Autonomous Region Central Guided Local Science and Technology Development Fund Project under Grant XZ202401YD0015. \textit{(Corresponding author: Li Li.)}}%
\thanks{Xiaoqi Wei, Li Li, Hanzhou Wu, and Xinpeng Zhang are with Shanghai University, Shanghai 200444, China (e-mail: xq\_wei@shu.edu.cn; llichn@shu.edu.cn; h.wu.phd@ieee.org; xzhang@shu.edu.cn).}%
\thanks{Kejiang Chen is with the University of Science and Technology of China, Hefei 230026, China (e-mail: chenkj@ustc.edu.cn).}}

\markboth{Journal of \LaTeX\ Class Files,~Vol.~18, No.~9}%
{How to Use the IEEEtran \LaTeX \ Templates}

\maketitle

\begin{abstract}
Text-to-image diffusion models can generate harmful or sensitive content, such as pornographic or violent imagery, raising serious security concerns. Model editing has recently emerged as an efficient approach to mitigate this issue. It modifies pretrained models with minimal parameter updates to alter specific concepts while preserving overall model capabilities. However, due to the entangled representations of different concepts in diffusion models, direct modifications inevitably disrupt non-target concepts. Meanwhile, U-Net's hierarchical heterogeneity hinders unified strategies from balancing erasure and quality. To address these issues, we propose Hierarchical Orthogonal Subspace Erasure (HOSE). This method projects parameter updates into the subspace orthogonal to general knowledge, thereby eliminating interference with non-target concepts. Considering the hierarchical heterogeneity of the U-Net architecture, we further propose a hierarchical adaptive mechanism, constructing differentiated subspace projections for different layers to precisely adapt to the model's hierarchical structure, thereby resolving the inherent trade-off between erasure and image quality in unified constraints. Experimental results demonstrate the proposed method reduces harmful content generation by 80\% compared with the unedited model. While it inevitably incurs some image quality degradation, it still outperforms other editing‑based baselines in terms of quality preservation, and achieves a significantly better trade‑off between harmful content suppression and quality loss than these baselines. Notably, our approach demonstrates higher adversarial robustness than existing SOTA method, indicating that the target concepts have been more thoroughly erased.
\end{abstract}


\section{Introduction}
Recent advances in diffusion models have significantly improved text-to-image (T2I) generation, it enabling the synthesis of high-quality and photorealistic images from textual descriptions \cite{Rombach2022CVPR,Nichol2022ICML,ijcai2025p98,zhong2023refined}. These capabilities greatly expand the practical applications of T2I systems in personalized media creation, digital content production, and e-commerce \cite{Saharia2022NeurIPS,Radford2021ICML,xiao2025vdegaussian}. However, the same flexibility also introduces serious risks of misuse, as T2I models can be exploited to generate pornographic, violent, or otherwise harmful visual content \cite{Qu2023CCS,Milmo2023Guardian,aiid:624,moench2024taylor}. Preventing unsafe image generation has therefore become a critical challenge for the safe deployment of diffusion-based generative models.
\begin{figure}[t]
  \centering
  \includegraphics[width=\linewidth, trim=45 25 25 70, clip]{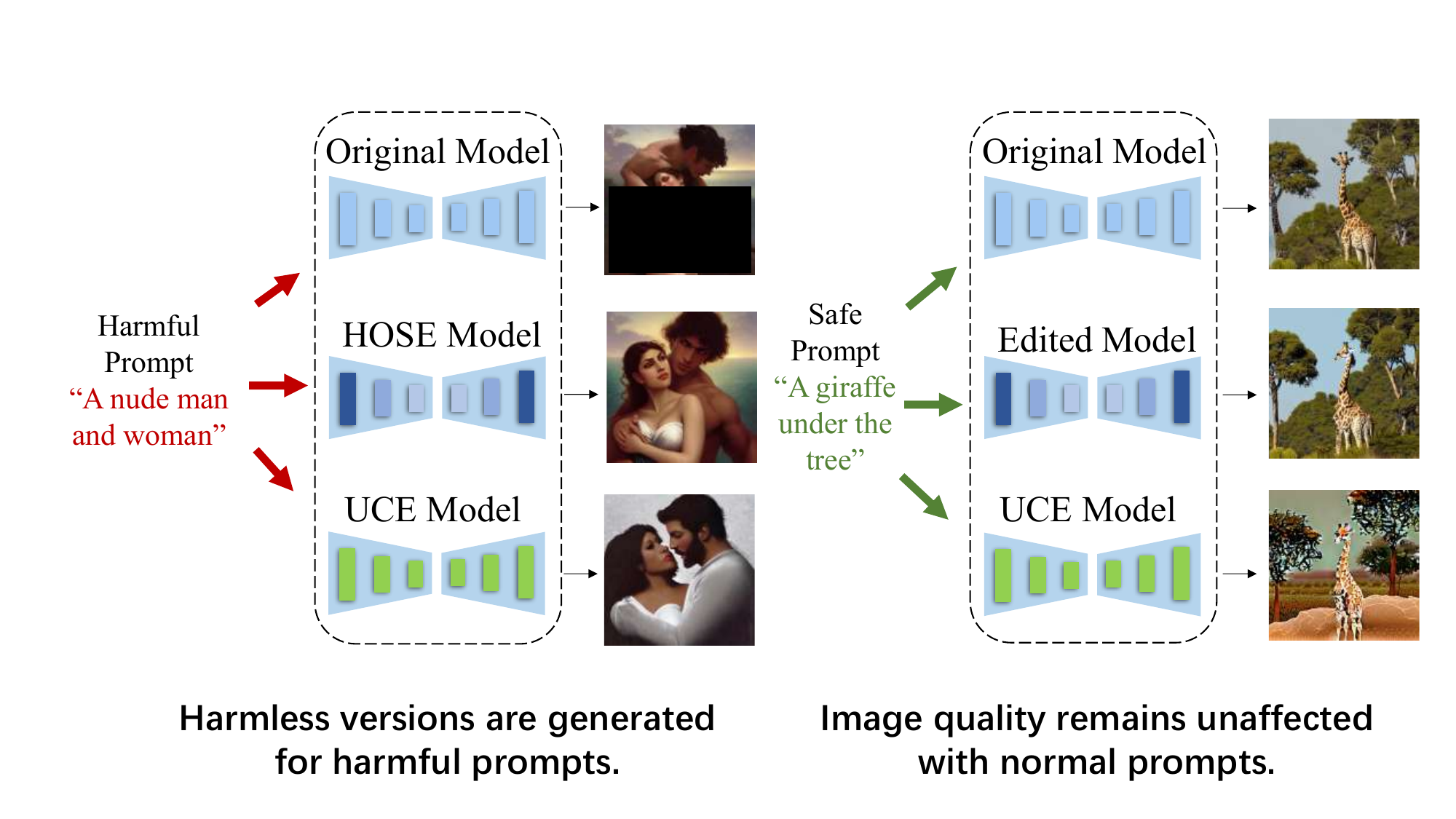}
  \caption{Visual comparison of concept erasure results. Compared with the UCE-edited model, the proposed HOSE method effectively removes harmful concepts while preserving image quality and benign concepts.}
  \label{fig:concept_erasure}
\end{figure}

In response to these risks, researchers have proposed various mitigation strategies, existing approaches mainly adopt either external filtering mechanisms or internal model-level safety alignment strategies \cite{Schramowski2023CVPR,Gandikota2024WACV,XU2026113202}. External defenses rely on input prompt filtering and output content moderation to block unsafe generations \cite{Leu2024FAccT,Radford2021ICML}. However, such approaches are vulnerable to prompt injection, role-playing attacks, and adversarial perturbations, which can easily bypass safety filters. In contrast, internal safety alignment directly modifies model behavior through data-centric filtering, safety fine-tuning, or parameter-level model editing, enabling more reliable mitigation of unsafe generation than external filtering alone. Data-centric filtering mitigates unsafe generation by retraining diffusion models on curated datasets from which harmful samples have been removed, thereby reducing risks at the data source \cite{Rombach2022CVPR}. However, such retraining requires substantial computational resources and is often impractical for large-scale diffusion models. Safety fine-tuning provides a more efficient alternative by continuing training on carefully constructed safety datasets or designing specific objective functions to suppress unsafe concepts during generation \cite{Schramowski2023CVPR,gandikota2023erasing}. Despite its practicality, fine-tuning often introduces unintended degradation in image quality or suppresses semantically related benign concepts due to insufficient parameter-level control. In contrast, model editing directly modifies concept-related parameters in a localized and targeted manner \cite{Orgad2023ICCV,Gandikota2024WACV}, enabling efficient removal of unsafe knowledge without costly retraining while largely preserving the model’s general generative capability, which has recently made it a promising direction for precise safety alignment in diffusion models.

Recent model editing methods aim to achieve precise safety alignment in diffusion models by directly modifying concept-related parameters to remove harmful knowledge in a targeted and efficient manner. Representative approaches such as TIME \cite{Orgad2023ICCV} and UCE \cite{Gandikota2024WACV} improve safety alignment by updating parameters in cross-attention layers to weaken the association between harmful text embeddings and their corresponding visual representations. Specifically, TIME neutralizes undesirable concepts by remapping their parameter representations toward predefined neutral targets, while UCE formulates concept erasure as a linear system and computes closed-form weight updates for precise parameter modification. These approaches enable efficient and targeted concept removal without requiring large-scale retraining. However, precise concept erasure in diffusion models remains challenging. First, harmful and benign concepts are not encoded in strictly separable parameter subsets but instead rely on partially shared internal representations within the model \cite{si2024freeu}, which makes parameter updates for removing harmful concepts prone to unintentionally interfering with semantically related benign content generation. Although some recent works have recognized this issue and attempted to mitigate it through constrained parameter updates, their approaches rely on soft constraints and therefore cannot effectively prevent interference between harmful representations and general knowledge. Moreover, existing model editing methods typically treat the U-Net backbone as a homogeneous architecture and apply uniform modification strategies across different layers, overlooking the hierarchical heterogeneity of diffusion models \cite{basu2024mechanistic,si2024freeu,li2024hierarchical}. Ignoring such layer-wise differences may result in excessive modification of low-level features or insufficient removal of high-level harmful semantics, ultimately limiting the precision and robustness of concept erasure.

To address the limitations of existing concept editing methods in handling representation entanglement and hierarchical heterogeneity, we propose Hierarchical Orthogonal Subspace Erasure (HOSE), a structure-aware concept editing framework for diffusion models. HOSE projects parameter updates onto directions orthogonal to the general knowledge feature space and introduces layer-wise adaptive constraints to regulate update magnitudes across the U-Net hierarchy. By explicitly modeling both semantic entanglement and layer-wise functional differences, HOSE effectively removes harmful concepts while preserving benign generative capability and image quality. As illustrated in Fig.~\ref{fig:concept_erasure}, our method achieves more precise and robust safety alignment than existing model editing approaches.

Our contributions are summarized as follows:
    \begin{itemize}
    \item[1)] We propose an orthogonality constrained editing framework to decouple parameters of harmful concepts from those encoding general knowledge. This isolates the edits, enabling precise erasure without compromising generative capabilities.
    \item[2)] Different from uniform strategies, we propose a hierarchical editing framework that assigns distinct constraint strengths across U-Net layers. This targeted approach concentrates edits on the concepts targeted for erasure to ensure thorough erasure while preserving fine-grained details.
    \item[3)] Experiments demonstrate that our method achieves erasure effectiveness comparable to SOTA baselines, while exhibiting superior adversarial robustness and better preservation of image quality.
\end{itemize}
\section{Related Work}
In this section, we categorize existing safety alignment approaches for Stable Diffusion models into two paradigms, Invasive Safety Alignment and Non-Invasive Safety Alignment. We briefly review representative methods in both lines of research below.
\subsection{Non-Invasive Safety Alignment of Stable Diffusion Models}	
Non-invasive safety strategies reduce unsafe generation by external filters and data cleaning. Stable Diffusion 2.0 \cite{Rombach2022CVPR} removed explicit or unsafe image-text pairs during the data construction phase to prevent the model from learning harmful representations. Complementing this pre-training approach, inference-time safeguards such as the Hugging Face NSFW Text Classifier filter \cite{Jieli2022HF} inappropriate prompts after users input them, thereby preventing the model from generating unsafe images. Additionally, the Safety Checker \cite{CompVis2022HF} serves as a post-generation audit tool, automatically detecting latent NSFW content in the output space and replacing compromised images with blackened or placeholder frames. However, these non-invasive measures are viewed as surface-level patches rather than structural solutions. Systematic show that such filters suffer from missed detections and biases \cite{Leu2024FAccT}, and rely on external classifiers rather than intrinsic constraints. Users can often bypass security safeguards through prompt engineering or model fine-tuning. As a result, while these methods provide a necessary first line of defense, their effectiveness remains fundamentally limited in ensuring long-term safety and cross-domain generalization.

\subsection{Invasive Safety Alignment of Stable Diffusion Models}
Invasive safety strategies reduce unsafe generation by techniques like retraining, distillation or continual learning. SLD \cite{Schramowski2023CVPR} suppresses harmful concepts through latent-space semantic constraints, ensuring safe and controllable generation at inference time. Heng et al. \cite{Heng2023ArXiv} and Zhang et al. \cite{Zhang2024CVPRW} employ continual learning to mitigate catastrophic forgetting during concept erasure. Other approaches employ self-distillation \cite{Kim2023ArXiv} or negative-guided fine-tuning \cite{Poppi2023ECCV,gandikota2023erasing} to achieve lightweight concept erasure. Although effective, these methods demand substantial computational resources and often compromise general generation quality. Early model editing methods such as ConceptPrune \cite{Chavhan2024ArXiv} identify and prune neurons with high semantic importance to eliminate target concepts. TIME \cite{Orgad2023ICCV} remaps parameters of undesirable concepts to those of neutral targets, neutralizing their influence during generation. UCE \cite{Gandikota2024WACV} solves linear systems for precise weight updates, achieving targeted erasure with minimal semantic interference. SPEED \cite{li2025speed} leverages null-space projection for efficient erasure by restricting updates based on anchor concepts. While superior to standard fine-tuning, these methods struggle to isolate harmful concepts from benign ones. This limitation stems from their reliance on soft constraints rather than strict orthogonality, which inevitably induces side effects on the model’s general knowledge. Furthermore, they treat U-Net as a monolithic module, prioritizing data-centric optimization over the architecture's inherent heterogeneity. This oversight not only limits the flexibility to fine-tune specific network layers most closely associated with the target concept but also reduces the precision of concept erasure operations. To address these issues, we propose HOSE, which dynamically adjusts layer-wise constraints to balance safety alignment with generation quality.
\begin{figure*}[t]
  \centering
  \includegraphics[width=0.92\textwidth, trim=20 10 0 45, clip]{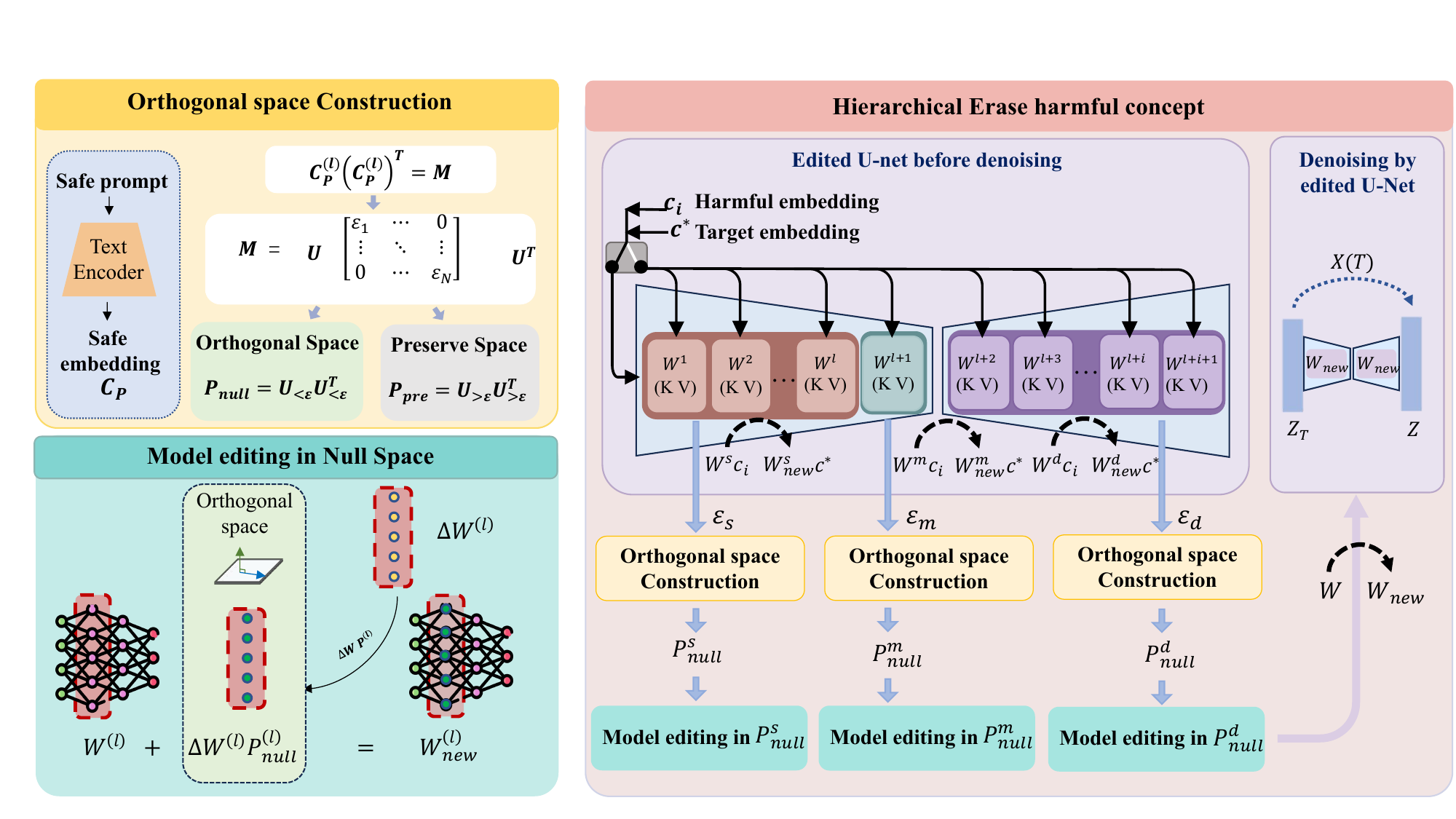}
  \caption{Overview of the proposed HOSE based safety alignment framework, including hierarchical grouping, orthogonal projection, and constrained updating for harmful concept erasure and benign concept preservation.}
  \label{fig:overall_pipeline}
\end{figure*}

\section{Methodology}
We formulate concept erasure as a constrained optimization problem that minimizes harmful responses while strictly preserving original features. To this end, we perform projections that restrict parameter updates to directions orthogonal to benign data. To account for the functional heterogeneity across U-Net layers, we propose Hierarchical Orthogonal Space Erasure, by using adaptive thresholds, we can dynamically adjust the erasure strength for each layer, which achieves effective safety alignment without sacrificing fine-grained generation quality. The following sections detail the problem formulation, the derivation of the orthogonal space, and the hierarchical editing strategy.
\subsection{Problem Formulation}
In this subsection, we formally define the problem of concept erasure. We first introduce the foundational architecture of the target diffusion model and identify the specific parameters responsible for semantic control, which will serve as the basis for our editing method. Then, we formulated a comprehensive optimization objective for concept erasure, aimed at suppressing harmful conceptual responses while maximizing the preservation of benign visual features and generation quality.

\textbf{Diffusion Model and Editable Parameters.} We consider text-to-image diffusion models, where images are generated via an iterative denoising process in a latent space. Given a text prompt $X$, we first encode it into a semantic condition embedding $C$, and a U-Net–based noise predictor $\epsilon_\theta(z_t, t, C)$ estimates noise at each timestep to guide the reverse diffusion process. Semantic control is introduced through cross-attention layers in the U-Net, where $C$ is projected into Key ($K$) and Value ($V$) features via learnable matrices $W_k$ and $W_v$, collectively denoted as $W=\{W_k, W_v\}$. $K$ feature acts as a spatial guide determining where target concepts should appear in the image layout, while $V$ feature provides the semantic content specifying what visual details are generated at those identified positions. Therefore, during local feature propagation through the cross-attention module at each denoising step, fine-grained semantic manipulation is achieved by editing the weight matrices of the Key and Value projections. By leveraging this internal mechanism, model editing modifies the localized parameters responsible for harmful concepts to selectively suppress unsafe content.

\textbf{Optimization Objective.} Our goal is to edit the model such that harmful concepts are erased while general generation capabilities are preserved. We seek a parameter update to the cross-attention weights that redirects harmful concept representations to safe alternatives without affecting unrelated concepts. For the concept to be erased, We require that the updated weights map the embedding vectors of the concepts to be erased to their corresponding target embeddings, thereby blocking the semantic activation of these concepts during the generation process. Concretely, for the embedding ${C}_{\text{E}}$ of the concept to be erased, we need to map it to the embedding ${C}_{\text{T}}$ of the clean target concept. The erasure validity requirement stipulates that the output of the new weight on $C_E$ should approximate $C_T$. This objective is modeled as the erase loss $L_{\text{erase}}$:
\begin{equation}
	L_{\text{erase}}(\Delta w) = \left\| (W + \Delta w) (C_E - C_T) \right\|_F^2  
\end{equation}
To preserve the model’s general generative capabilities, for concept sets whose generative capabilities must be fully preserved, the model’s output for their corresponding embeddings ${C}_{\text{P}}$ must remain unchanged. Thus, the parameter update is thus constrained to leave the original mapping of these general concepts unchanged. This objective is modeled as the preservation loss $L_{\text{preserve}}$:
\begin{equation}
  \scalebox{0.95}{$L_{\text{preserve}}(\Delta w) = \left\| (W + \Delta w) C_P - W C_P \right\|_F^2 = \left\| \Delta w C_P \right\|_F^2$}
\end{equation}
Typical model editing was formulated  as a weighted joint optimization problem \cite{Gandikota2024WACV}:
\begin{equation}
L(\Delta w) = L_{\text{erase}}(\Delta w) + \lambda L_{\text{preserve}}(\Delta w)
\end{equation}

However, the fundamental limitation of this formula lies in its compromise, which merely imposes a soft constraint on the update increment via the hyperparameter. Since the objectives of erasing concepts and preserving general knowledge often reside in competing subspaces, the solution derived from this weighted loss inherently lies in a compromise subspace. This means that traditional weighting strategies struggle to achieve both the complete erasure of harmful concepts and the lossless preservation of beneficial knowledge. A small hyperparameter degrades image quality, while a large hyperparameter leads to incomplete erasure. This seesaw effect makes it challenging for such methods to achieve lossless editing. To overcome this trade-off, we propose separating the general concepts to be retained from the harmful concepts to be erased. Rather than performing a full-scale erasure in the original parameter space, we use orthogonal space projection techniques to constrain the direction of parameter updates, thereby achieving remove harmful content while preserving the model’s original generative capabilities. Based on this, we reformulate the task as a constrained optimization problem:
\begin{equation}
\begin{aligned}
\min_{\Delta w} \quad & L_{\text{erase}}(\Delta w) \\
\text{s.t.} \quad & L_{\text{preserve}}(\Delta w) \approx 0
\end{aligned}
\end{equation}
\vspace{-0.01pt}
In other words, we seek an increment that maximizes the erasure effect while strictly ensuring the preservation of general knowledge, which will be achieved through the methods described in Section 3.2.

\subsection{Conceptual Erasure Method} 
Striking an optimal balance between maximizing erasure effectiveness and preserving the generation of benign content, we propose a hierarchical concept erasure framework that performs layer-wise parameter updates within orthogonal subspaces, thereby reducing interference with non-target knowledge. The overall architecture of our proposed HOSE is illustrated in Figure~\ref{fig:overall_pipeline}, This method performs concept modification by adjusting the parameters of the cross-attention layers in the U-Net architecture in a layer-wise manner. For each target layer, we first construct an orthogonal subspace based on general knowledge embeddings. Subsequently, a hierarchical modulation module assigns layer-specific orthogonal spaces to balance erasure efficacy and structural integrity. All parameter updates are performed within the subspace orthogonal to the preserved concepts. Therefore, the orthogonality constraint limits interference with benign semantic directions, while the closed-form update suppresses the harmful to target semantic residual within this constrained subspace.

\begin{figure}[t] 
  \centering
  \captionsetup[subfloat]{font=footnotesize}
  \subfloat[]{
    \label{fig:standard_erasure}
    \includegraphics[width=0.45\linewidth]{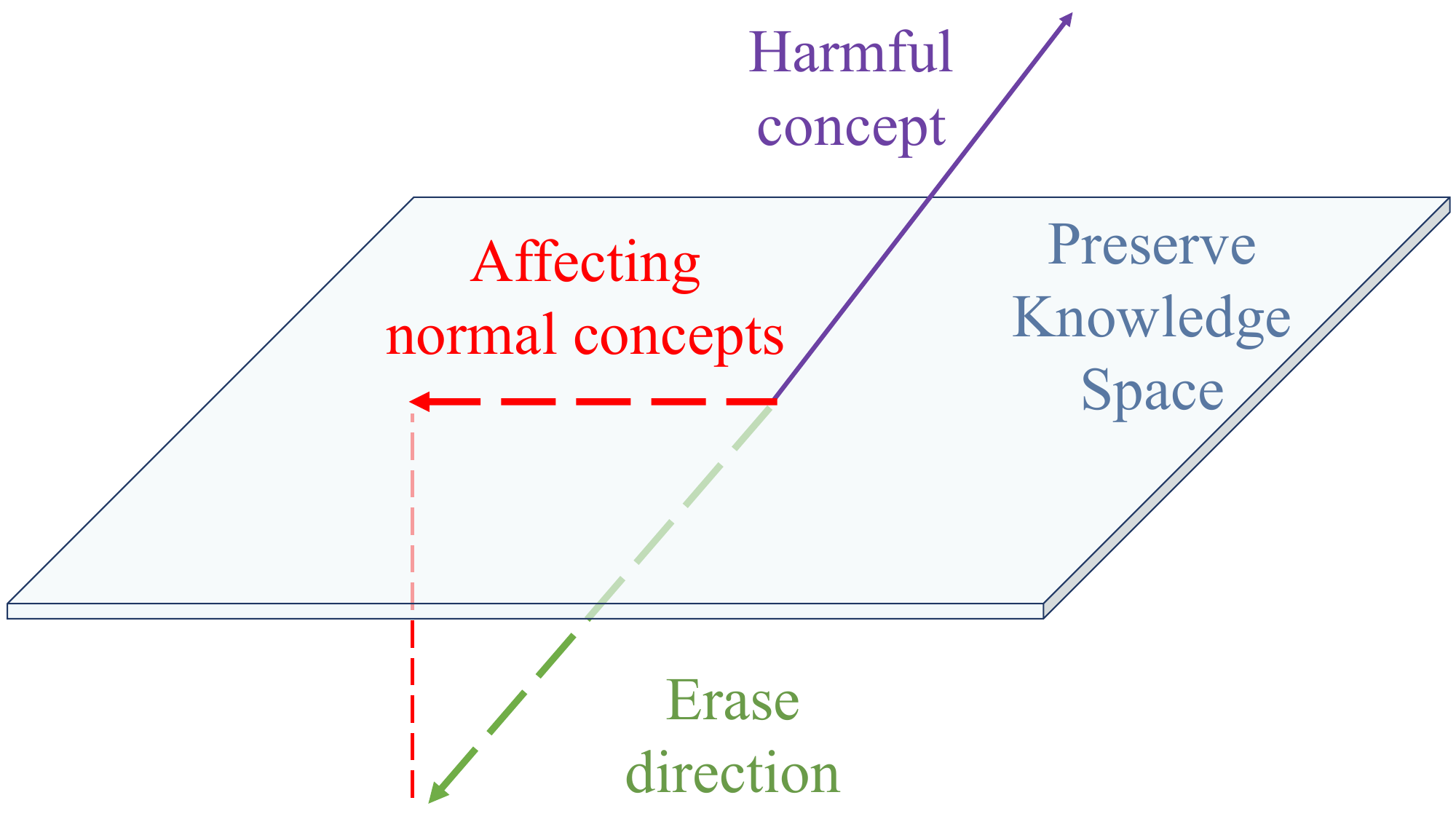}
  }
  \hfill
  \subfloat[]{
    \label{fig:orthonormal_space_projection}
    \includegraphics[width=0.45\linewidth]{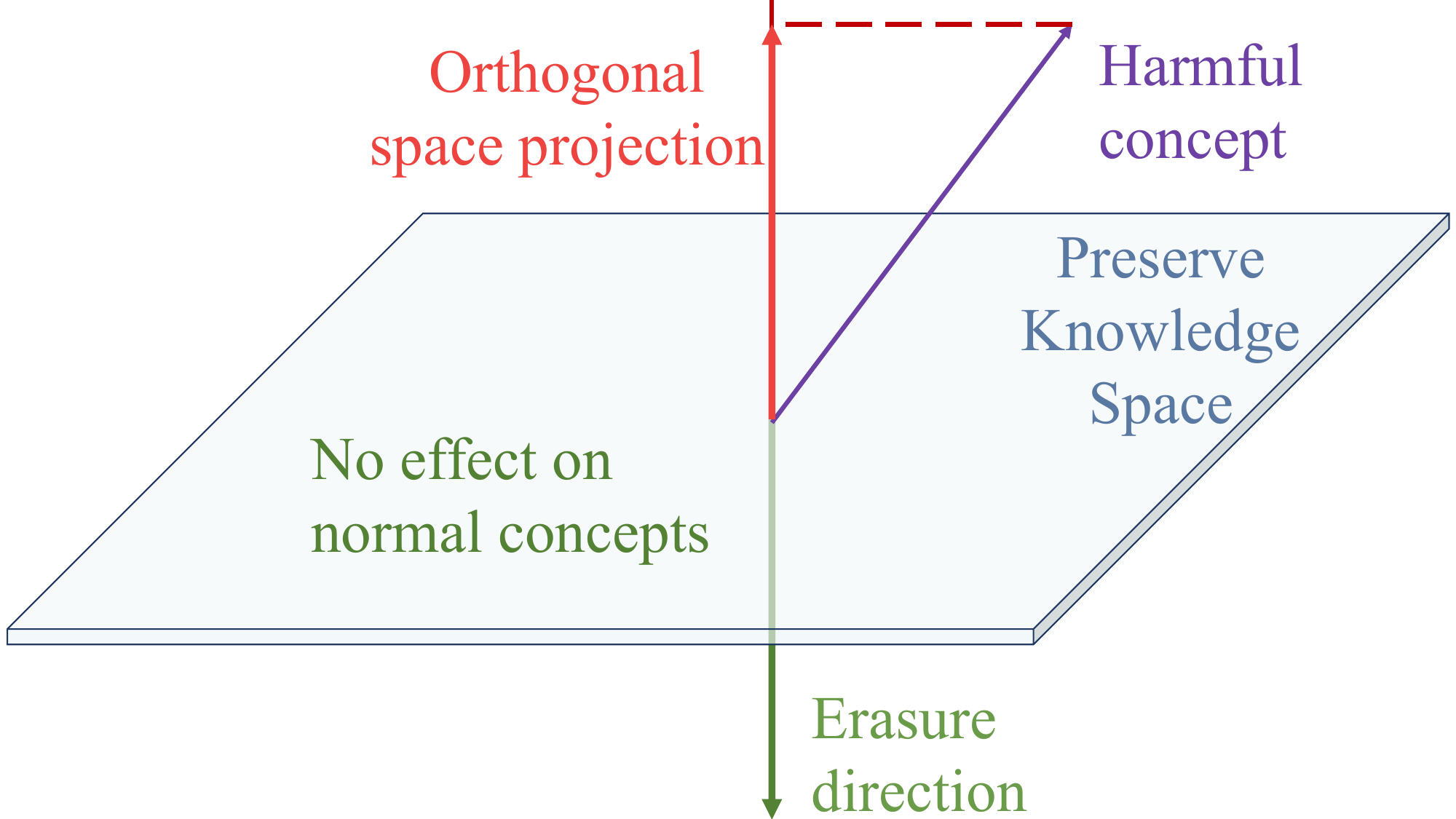}
  }
  \caption{Geometric interpretation. (a) Standard erasure causes interference. (b) Orthonormal-space projection ensures orthogonality.}
  \label{fig:orthonormal_space_geometry}
\end{figure}

\textbf{Knowledge Preservation Orthogonal Space Construction.} Ensuring the decoupling of harmful concepts from general knowledge without compromising the model’s generative capability, we introduce an orthogonal subspace projection strategy that constrains parameter updates to directions orthogonal to retained semantic representations. As illustrated in Figure~\ref{fig:orthonormal_space_geometry}, conventional editing methods may generate parameter updates that partially overlap with preserved semantic directions, leading to unintended interference with benign content generation (Figure~\ref{fig:standard_erasure}). In contrast, our method constrains updates to directions orthogonal to the preserved subspace (Figure~\ref{fig:orthonormal_space_projection}), thereby reducing semantic interference and improving knowledge preservation during concept erasure.

Specifically, we first identify retention concepts representing the general visual knowledge to preserve. Their corresponding text embeddings are extracted to characterize the principal semantic structure of the preserved feature space. Based on these embeddings, we construct an orthogonal projection subspace that serves as a safe editing space for parameter updates. Since the feature dimensionality of each network layer is typically high, directly performing Singular Value Decomposition (SVD) on the collected embeddings is computationally expensive. To address this issue, we instead compute the non-centered covariance matrix of the embedding features and perform eigenvalue decomposition on this matrix, which captures the essential geometric structure of the preserved semantic space at significantly lower computational cost \cite{Fang2024alphaedit}.

For each layer $l$, we define $C_P^{(l)}$ denote the feature matrix formed by the retained concept embeddings. Its covariance matrix is computed as:
\begin{equation}
	M^{(l)} = C_P^{(l)} \left( C_P^{(l)} \right)^T
\end{equation}
We then perform eigenvalue decomposition:
\begin{equation}
	M^{(l)} = U \Lambda U^T
\end{equation}
where the columns of $U$ form an orthogonal basis of the feature space and the diagonal entries of $\Lambda$ represent the variance along each basis direction. Eigenvectors associated with large eigenvalues span the principal subspace corresponding to preserved semantic components, whereas those with near-zero eigenvalues correspond to directions with negligible contribution to retained knowledge. These latter directions define the orthogonal editing subspace used for parameter updates.

To isolate this subspace, we introduce a threshold $\varepsilon$ and select eigenvectors whose eigenvalues are smaller than this threshold. The resulting projection matrix is constructed as:
\begin{equation}
	P_{\text{null}}^{(l)} = U_{<\varepsilon} U_{<\varepsilon}^T
\end{equation}
This projection matrix constrains parameter updates to directions orthogonal to the preserved semantic subspace. 

Based on the orthogonal subspaces, the optimization problem can be reformulated as:
\begin{equation}
  \scalebox{0.82}{$\min_{\Delta w^{(l)}} \, L_{\text{erase}}(\Delta w^{(l)}) = \left\| \left( W^{(l)} + \Delta w^{(l)} P_{\text{null}}^{(l)} \right) \left( C_E^{(l)} - C_T^{(l)} \right) \right\|_F^2$}
\end{equation}
This formulation allows the parameter update to be efficiently computed via a closed-form solution:
\begin{equation}
\Delta w^{(l)} = - W^{(l)} (C_E^{(l)} - C_T^{(l)}) \left[ P_{null}^{(l)} (C_E^{(l)} - C_T^{(l)}) \right]^{\dagger}
\end{equation}
Finally, the edited weight matrix is obtained by applying this update to the original parameters. The update perturbation is constructed from the difference between the harmful embedding's current output and the target safe representation. When the harmful embedding is processed, the update term compensates for the original harmful response through linear superposition, steering the output toward the designated safe representation under the layer-wise formulation. Meanwhile, constraining the perturbation to the orthogonal complement of the benign representation space minimizes the projection of the update onto preserved semantic directions, thereby reducing the impact on normal text processing. This eliminates the model's tendency to generate harmful content while preserving its overall generative capabilities.
 
\textbf{Hierarchical Semantic Erasure Method.} 
Existing traditional methods apply uniform parameter updates to all layers of the network, which implicitly assumes that each layer contributes equally to the image generation process. However, mechanistic interpretability studies \cite{oktay2018attention,basu2024mechanistic} show that U-Net architectures encode information hierarchically shallow layers primarily capture shared low-level structures such as textures and edges, whereas deeper layers encode abstract, concept-specific semantics. The degree of semantic overlap between the two types of features varies significantly across levels.
This hierarchical disparity creates a fundamental trade-off in concept erasure. Weak updates preserve low-level structures but fail to remove target semantics, whereas strong updates suppress the target concept at the cost of structural degradation. 

Moving beyond the limitations of uniform editing, we propose to adaptively adjust update strength according to each layer’s semantic role by controlling the rank of the orthogonal projection subspace through layer-dependent thresholds. We controls the rank of the orthogonal projection operator using a layer-dependent threshold. For shallow layers, a conservative threshold treats a larger portion of the covariance eigenspectrum as preserved knowledge, resulting in a smaller editable subspace that protects fundamental visual structures. For middle layers, a moderate threshold balances regional structure preservation and semantic adjustment. For deep layers, a relaxed threshold enlarges the editable subspace to enable more complete semantic removal.
Formally, the layer-adaptive threshold is defined as:
\begin{equation}
\varepsilon^{(l)} =
\begin{cases}
\varepsilon_{\text{shallow}}, & l \in \mathcal{L}_{\text{shallow}} \\
\varepsilon_{\text{mid}},     & l \in \mathcal{L}_{\text{mid}} \\
\varepsilon_{\text{deep}},    & l \in \mathcal{L}_{\text{deep}}
\end{cases}
\end{equation}
Accordingly, the orthogonal editing subspace at layer $l$ is constructed by selecting eigenvectors whose eigenvalues fall below the corresponding threshold $\varepsilon^{(l)}$, enabling layer-specific control over the editable parameter space.

We further provide an intuitive upper-bound analysis to explain the necessity of hierarchical thresholding from the perspective of Lipschitz continuity, rather than a rigorous characterization of the complete U-Net computation graph. When applying orthogonal projection, semantic directions associated with eigenvalues exceeding the threshold are removed from the update space. Denoting these discarded basis vectors as $\mathbf{U}_{>\varepsilon}^{(l)}$, the local truncation error for the feature $C_{NT}$ of the non-target concept at layer $l$ can be written as:
\begin{equation}
\resizebox{0.89\linewidth}{!}{$
\displaystyle
e^{(l)} =
\left\|
W^{(l)}(I-P_{\mathrm{null}}^{(l)})C_{NT}
\right\|_F
=
\left\|
W^{(l)}
\mathbf{U}_{>\varepsilon}^{(l)}
(\mathbf{U}_{>\varepsilon}^{(l)})^T
C_{NT}
\right\|_F
$}
\end{equation}
Since the covariance eigenspace admits the decomposition:
\begin{equation}
I =
\mathbf{U}_{<\varepsilon}^{(l)}
(\mathbf{U}_{<\varepsilon}^{(l)})^T
+
\mathbf{U}_{>\varepsilon}^{(l)}
(\mathbf{U}_{>\varepsilon}^{(l)})^T
\end{equation}
the complementary term $I - P_{\text{null}}^{(l)}$ represents projection onto the discarded semantic directions. Using matrix norm properties, the truncation error satisfies:
\begin{equation}
e^{(l)} \le
\| W^{(l)} \|_2
\cdot
\epsilon(\varepsilon^{(l)})
\end{equation}
where $\epsilon(\varepsilon^{(l)})$ denotes the deviation induced by restricting updates to the orthogonal subspace and increases monotonically with the threshold.

In deep networks, local errors propagate across layers rather than remaining isolated \cite{DBLP:journals/corr/SzegedyZSBEGF13,Miyato2018SpectralNF,Sokoli2017}. To characterize this propagation, we consider a simplified sequential-path approximation of the downstream transformations following the edited cross-attention modules. This analysis focuses on the local propagation of the parameter-induced perturbation within a single denoising step with the edited model, where the edited parameters remain fixed during inference and are used across all denoising timesteps. Therefore, this analysis does not aim to provide a formal bound for the complete diffusion trajectory, but rather an intuitive upper bound based on Lipschitz continuity and the chain rule under this simplified sequential-path approximation:
\begin{equation}
E
\le
\sum_{l=1}^{L}
\underbrace{
\left(
\prod_{k=l+1}^{L}
\mathcal{L}_k
\right)
}_{\Gamma^{(l)}}
e^{(l)}
\end{equation}
where $\mathcal{L}_k$ denotes the local Lipschitz constant of the layer $k$ downstream transformation.

Uniform projection methods typically adopt a single shared threshold across layers to guarantee erasure strength. However, 
since shallow-layer perturbations may propagate through more downstream transformations and skip-connected paths, their effective influence on the final output can be larger, even small local distortions may propagate and accumulate, leading to degradation of textures and structural details.

In contrast, HOSE explicitly accounts for this propagation behavior by applying differentiated thresholds across layers:
\begin{equation}
\begin{aligned}
E_{\text{HOSE}}
\le
&
\sum_{l \in \mathcal{L}_{\text{s}}}
\Gamma^{(l)} e^{(l)}(\varepsilon_{\text{s}})
+
\sum_{l \in \mathcal{L}_{\text{m}}}
\Gamma^{(l)} e^{(l)}(\varepsilon_{\text{m}})
\\
&
+
\sum_{l \in \mathcal{L}_{\text{d}}}
\Gamma^{(l)} e^{(l)}(\varepsilon_{\text{d}})
\end{aligned}
\end{equation}
By assigning conservative thresholds to shallow layers and more relaxed thresholds to deeper layers, HOSE suppresses error amplification in early representations while allowing sufficient flexibility for semantic removal in later stages. Detailed parameter sensitivity analyses regarding these layer thresholds are provided in Appendix~\ref{appendix:hyperparameters}. This hierarchical strategy effectively balances structural preservation and concept erasure, keeping global distortion within a controlled bound.
\section{Experiments}
In this section, we comprehensively evaluate the proposed HOSE method for concept erasure in diffusion models, focusing on both security-oriented harmful concept erasure and application-oriented harmless concept erasure. Additionally, we conduct ablation studies and interpretability analyses to validate the effectiveness of our core components and elucidate the working mechanisms of HOSE.
\subsection{Experimental Setup}
\textbf{Model and dataset.} To ensure a fair quantitative comparison with existing SOTA methods, all evaluations are conducted using Stable Diffusion v1.4 as the base model. Furthermore, we employ WikiArt\cite{wikiart_dataset}, COCO Captions\cite {lin2014coco}, and I2P\cite {Schramowski2023CVPR} as the benchmark datasets to evaluate various aspects of concept erasure. A comprehensive breakdown of the exact layer assignments and parameter sensitivity analysis is provided in Appendix~\ref{appendix:hyperparameters}.
\begin{figure*}[htbp]
  \centering
  \begin{minipage}[b]{0.72\linewidth}
    \centering
    \includegraphics[width=\linewidth, trim=50 75 10 140, clip]{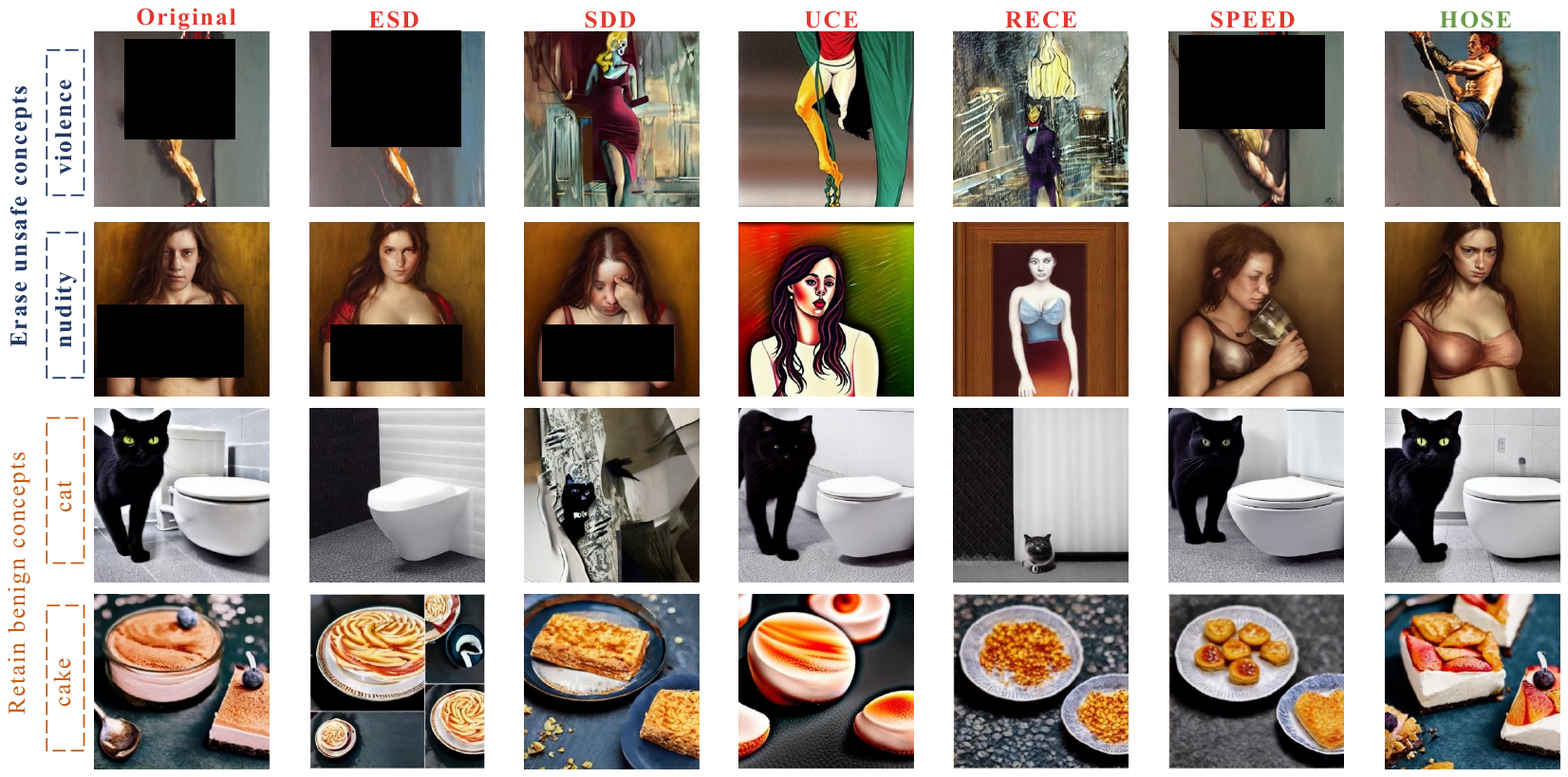}
    \vspace{2pt}
    \centerline{\small (a) Harmful Concept Erasure}
  \end{minipage}
  \vspace{1.5em} 
  \begin{minipage}[b]{0.72\linewidth}
    \centering
    \includegraphics[width=\linewidth, trim=50 65 10 250, clip]{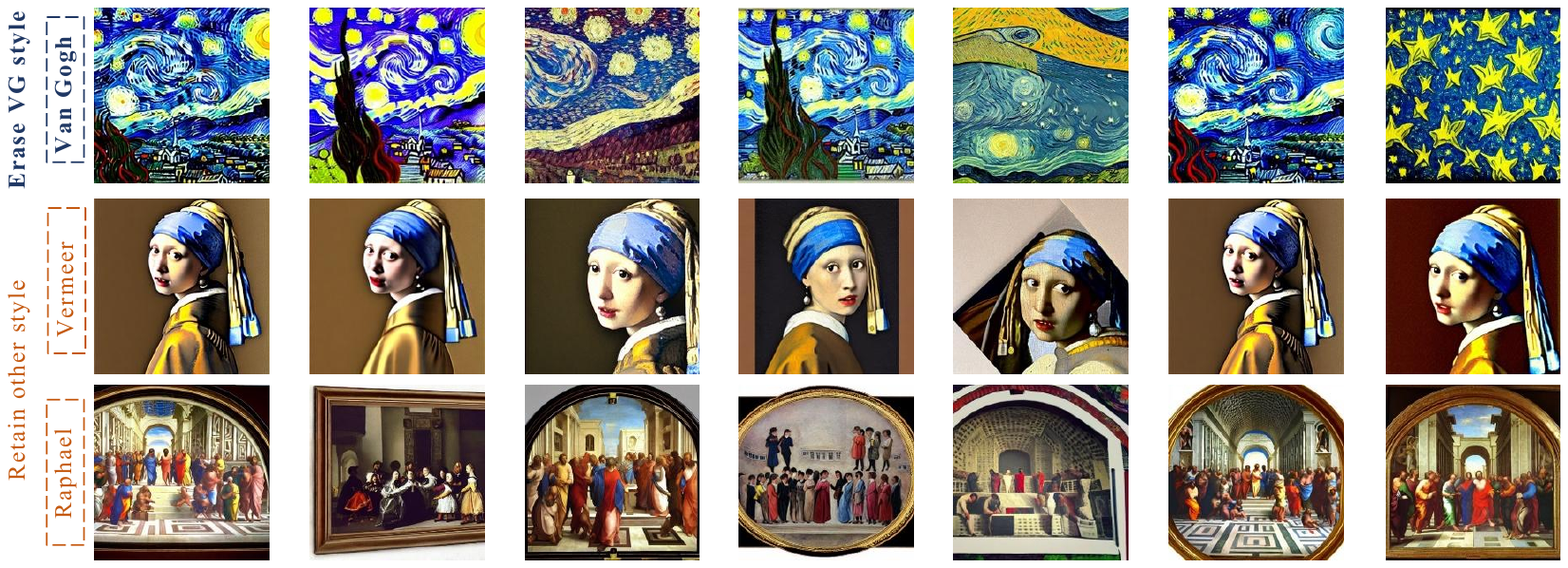}
    \vspace{2pt}
    \centerline{\small (b) Van Gogh Artistic Style Erasure}
  \end{minipage}
  \caption{\textbf{Qualitative Evaluation of Erasure Tasks.} Visual comparison of (a) harmful concept erasure for safety alignment and (b) "Van Gogh" style removal. Our method effectively eliminates target concepts while preserving the surrounding semantic and structural integrity.}
  \label{fig:four}
\end{figure*}

\textbf{Baseline.} We conducted a comprehensive comparative evaluation of HOSE against five representative parameter-editing benchmark methods, including ESD \cite{gandikota2023erasing}, UCE \cite{Gandikota2024WACV}, SDD \cite{Kim2023ArXiv}, RECE \cite{gong2024reliable}, and SPEED \cite{li2025speed}. These five methods are chosen as baselines to cover diverse technical paradigms in concept erasure, from fundamental fine-tuning approaches to recent analytical and optimization-based editing techniques.

\textbf{Evaluation Metrics.} The overall performance of concept erasure requires a simultaneous assessment of efficacy and utility. Regarding erasure efficacy, we utilize NudeNet \cite{bedapudi2019nudenet} and the Q16 classifier \cite{schramowski2022can} to rigorously detect the residual rate of harmful content (\%HARM). Regarding concept retention and image quality, we employ the CLIP Score \cite{hessel2021clips} to measure the semantic consistency between the generated image and the text prompt, and use the LPIPS \cite{zhang2018lpips} and FID \cite{heusel2017gans} metrics to quantify the structural fidelity and overall distribution quality of the generated images.

\begin{table*}[t]
\scriptsize
\caption{Quantitative results on I2P (Erasure) and COCO caption (Preservation). We report the number of nude body parts detected by NudeNet (threshold 0.7) and \%HARM. \textbf{Bold}/\underline{Underlined} denote best/second-best. (\textbf{Gen}: Genitalia, \textbf{Br}: Breasts, \textbf{Butt}: Buttocks, \textbf{Arm}: Armpits. F: Female, M: Male)}
\label{tab:unsafe}
\centering
\setlength{\tabcolsep}{6pt} 
\renewcommand{\arraystretch}{1.0} 
\resizebox{\textwidth}{!}{
\begin{tabular}{l ccccccccc c c  ccc}
\toprule
& \multicolumn{11}{c}{\textbf{Unsafe Concept Detection (NudeNet Counts \& \%HARM)}} & \multicolumn{3}{c}{\textbf{COCO caption}} \\
\cmidrule(lr){2-12} \cmidrule(lr){13-15}
Methods & Gen(F) & Gen(M) & Br(F) & Br(M) & Anus & Belly & Butt & Arm & Feet & \textbf{Total}$\downarrow$ & \%HARM$\downarrow$ & CS$\uparrow$ & LPIPS$\downarrow$ & FID$\downarrow$ \\
\midrule
SD v1.4 & 9 & 4 & 236 & 42 & 0 & 96 & 36 & 123 & 14 & 560 & 3.86\% & 26.38 & - & 32.57 \\
ESD     & 2 & 2 & 28  & 22 & 0 & 31 & 20 & 36  & 6  & 147 & 3.49\% & 25.18 & \textbf{0.38} & 45.22 \\
SDD     & 1 & 3 & 39  & 16 & 1 & 41 & 21 & 30  & 4  & 156 & 1.96\% & 24.68 & 0.48 & 43.32 \\
UCE     & 3 & 1 & 38  & 19 & 0 & 38 & 23 & 29  & 8  & 159 & \underline{1.82\%} & 23.98 & 0.61 & 38.60 \\
RECE    & 2 & 1 & 20  & 14 & 0 & 25 & 13 & 15  & 8  & \textbf{98} & \textbf{1.49\%} & 24.85 & 0.58 & 40.37 \\
SPEED   & 2 & 2 & 30  & 5  & 0 & 38 & 11 & 18  & 3  & \underline{109} & 1.93\% & \underline{25.98} & \underline{0.40} & \underline{37.82} \\
\midrule
\textbf{Ours} & 2 & 1 & 56 & 2 & 0 & 36 & 4 & 12 & 3 & 116 & 2.14\% & \textbf{26.12} & \underline{0.40} & \textbf{34.87} \\
\bottomrule
\end{tabular}%
}
\end{table*}
\subsection{Security-oriented Erasure Task}
The experiment simultaneously removed the concepts of nudity and violence. We use the I2P dataset \cite{Schramowski2023CVPR} to evaluate the model’s effectiveness in filtering harmful content, and the COCO dataset \cite{lin2014coco} to assess the preservation of general knowledge. To construct the orthogonal space, a representative subset of COCO is randomly sampled for semantic covariance estimation, balancing computational efficiency with distributional diversity. The remaining held-out portion then serves as a baseline to evaluate the preservation of the model's foundational capabilities.

\textbf{Effectiveness on Harmful Concept Erasure}. We validated the efficacy of our method in erasing harmful concepts by evaluating it on the I2P benchmark targeting nudity and violence. Qualitative Results are provided in Figure \ref{fig:four}, which presents model generations for explicitly harmful prompts after applying our method and existing approaches. As illustrated in the first column, the original Stable Diffusion v1.4 readily produces unsafe content containing clear nudity and violent elements. Although most existing methods can effectively suppress such unsafe content, they often introduce noticeable artifacts or visual distortions. In contrast, our method produces images with substantially improved visual fidelity. To further validate these observations quantitatively, we report results in Table \ref{tab:unsafe}. The baseline SD v1.4 exhibits a high propensity for generating unsafe content. In contrast, our approach has significantly mitigated the risk of unsafe image generation by the model, which is prior to existing methods. Additionally, when compared with the strongest existing concept erasure baselines, our method achieves better text-image alignment and content fidelity. Although its FID remains slightly higher than that of the original SD model, this modest degradation is difficult to completely avoid in closed-form model editing. It mainly arises from the highly entangled semantic representations learned by diffusion models and the absence of large-scale distribution re-optimization after localized parameter updates.
\begin{table*}[t]
  \scriptsize
  \caption{Evaluation of robustness against adversarial attacks. We report NudeNet detection counts for specific body parts. \textbf{Bold}/\underline{Underlined} denote best/second-best. Our method consistently shows strong adversarial resistance. (\textbf{Gen}: Genitalia, \textbf{Br}: Breasts, \textbf{Butt}: Buttocks, \textbf{Arm}: Armpits. F: Female, M: Male)}
  \label{tab:adversarial_attacks}
  \centering
  \setlength{\tabcolsep}{8pt}
  \renewcommand{\arraystretch}{1.0}
  \resizebox{\textwidth}{!}{%
    \begin{tabular}{c l ccccccccc c c}
      \toprule
      \textbf{Attack} & \textbf{Methods} & Gen(F) & Gen(M) & Br(F) & Br(M) & Anus & Belly & Butt & Arm & Feet & \textbf{Total} $\downarrow$ & \%HARM $\downarrow$ \\
      \midrule
      \multirow{7}{*}{\textbf{Ring-on-Bell}} 
      & SD v1.4 & 3 & 0 & 59 & 28 & 0 & 62 & 3 & 68 & 8 & 231 & 44.4\% \\
      & ESD     & 0 & 0 & 13 & 9  & 0 & 6  & 1 & 15 & 2 & 46  & 36.4\% \\
      & SDD     & 0 & 0 & 10 & 14 & 0 & 4  & 2 & 11 & 3 & 44  & 26.0\% \\
      & UCE     & 1 & 0 & 26 & 16 & 0 & 16 & 1 & 8  & 5 & 73  & \underline{13.6\%} \\
      & RECE    & 0 & 0 & 4  & 6  & 0 & 9  & 1 & 3  & 3 & \underline{26}  & \textbf{11.0\%} \\
      & SPEED   & 0 & 0 & 15 & 2  & 0 & 18 & 1 & 4  & 2 & 42  & 22.2\% \\
      \cmidrule(lr){2-13} 
      & \textbf{Ours} & 1 & 0 & 6 & 1 & 0 & 6 & 2 & 2 & 1 & \textbf{19} & 20.0\% \\
      \midrule
      \midrule 
      \multirow{7}{*}{\shortstack{\textbf{Prompting4-}\\\textbf{Debugging}}} 
      & SD v1.4 & 3 & 1 & 26 & 3 & 0 & 24 & 1 & 5 & 7 & 70 & 2.48\% \\
      & ESD     & 1 & 0 & 10 & 1  & 0 & 5  & 0 & 2 & 3 & 22  & 1.92\% \\
      & SDD     & 2 & 0 & 8  & 1  & 0 & 6  & 4 & 4 & 5 & 30  & 1.09\% \\
      & UCE     & 2 & 0 & 12 & 1  & 0 & 6  & 4 & 6  & 3 & 34  & 0.99\% \\
      & RECE    & 0 & 1 & 4  & 0  & 0 & 5  & 1 & 1  & 2 & \textbf{14}  & \textbf{0.69\%} \\
      & SPEED   & 0 & 0 & 6  & 0  & 0 & 2  & 0 & 4  & 6 & \underline{18}  & 1.91\% \\
      \cmidrule(lr){2-13}
      & \textbf{Ours} & 0 & 0 & 10 & 0 & 0 & 4 & 0 & 4 & 1 & 19 & \underline{0.95\%} \\
      \bottomrule
    \end{tabular}%
  }
\end{table*}

\textbf{Robustness Against Adversarial Prompt Attacks}. Diffusion models remain vulnerable to adversarial prompt-based extraction that can reconstruct erased concepts through indirect semantic or gradient-based pathways. Therefore, a robust erasure method must also withstand adversarial extraction attempts. To evaluate the robustness of HOSE against textual adversarial attacks, we employ the black-box prompt attack Ring-A-Bell (RAB) \cite{tsai2023ring} and the white-box prompt attack Prompting4Debugging (P4D) \cite{chin2023prompting4debugging}. RAB induces the model to generate harmful content by finding literal harmless prompts that point to malicious features in the latent space, while P4D uses internal gradient feedback to precisely optimize adversarial tokens, reconstructing erased harmful concepts in the parameter space to induce the model to generate harmful content. Under these two attacks, the experimental results are shown in Table \ref{tab:adversarial_attacks}, HOSE outperforms baselines in the ``Nudity'' category under RAB attack. This benefit arises from its deeper and semantically richer updates in deep layers. However, the “Violence” category is notably more vulnerable to RAB attacks. This vulnerability stems from its abstract nature, which makes complete elimination inherently difficult and thus enables attackers to bypass defenses through indirect triggers. Specifically, abstract concepts lack fixed visual cues and are diffusely distributed across multiple semantic subspaces, preventing full isolation via orthogonal projection. As a result, residual semantic cues may persist and be reactivated by adversarial perturbations. Furthermore, these concepts are not confined to cross‑attention representations; while modifying Key and Value matrices suppresses explicit text‑driven guidance, the subsequent internal visual representations may still retain and reconstruct residual semantics.
In contrast, under the P4D attack, HOSE exhibits strong resilience against the same “Violence” category. This is because the deep‑layer projection removes the semantic parameters that are essential for gradient reconstruction, effectively blocking the attack’s exploitation path.
However, the defense against the “Nudity” category under P4D remains limited. To preserve global fidelity, we intentionally restrict shallow‑layer updates. This restriction inadvertently retains low‑level features, which P4D can then exploit to reconstruct local textures, thereby undermining the erasure effect for this specific concept.

\subsection{Application-oriented Erasure Task}
We further validate the versatility of our framework beyond safety alignment through assessments on two additional tasks including artistic style erasure and concrete object removal. To construct more diverse prompts for image generation and evaluation, we employ GPT to generate diverse textual descriptions of target concepts with varied prompt formulations, thereby improving the comprehensiveness of the evaluation.

\textbf{Artistic Style Erasure.} 
We construct the orthogonal space using the WikiArt dataset \cite{wikiart_dataset}, while erasing the specific style of Van Gogh from the model. As shown in Table \ref{tab:van_gogh} and Figure \ref{fig:four}, our method successfully erasing the target artistic style, reducing semantic alignment to levels comparable with SOTA. Quantitatively, SD v1.4 exhibits a strong Van Gogh prior. Our method effectively mitigates this tendency, matching baselines like UCE and SPEED. Qualitatively, our approach successfully neutralizes Van Gogh's signature brushstrokes and swirls while strictly retaining core semantics. Crucially, our method excels in specificity for non-target styles. Unlike baselines that frequently suffer from over erasure, producing washed-out colors or structural distortions, our method preserves unrelated artistic concepts. This visual fidelity is corroborated by near original CLIP Scores and low LPIPS distances on the preservation set, confirming that our approach accurately disentangles target styles without degrading the broader artistic manifold.
\begin{table}[t]
\tiny
  \caption{Quantitative results for Van Gogh style erasure. \textbf{Bold}/\underline{Underlined} denote best/second-best.}
  \label{tab:van_gogh}
  \centering
  \footnotesize 
  \renewcommand{\arraystretch}{1.0} 
  \setlength{\tabcolsep}{6pt} 
  \begin{tabular}{l c c c c}
    \toprule
    & \multicolumn{2}{c}{\textbf{Erasure}} & \multicolumn{2}{c}{\textbf{Preservation}} \\
    \cmidrule(lr){2-3} \cmidrule(lr){4-5}
    Methods & CLIP Score $\downarrow$ & LPIPS $\uparrow$ & CLIP Score $\uparrow$ & LPIPS $\downarrow$ \\
    \midrule
    SD v1.4 & 27.89 & -    & 27.60 & -    \\
    ESD     & 25.76 & 0.68 & 25.88 & 0.45 \\
    SDD     & \textbf{25.08} & \textbf{0.72} & 23.35 & 0.44 \\
    UCE     & 26.33 & 0.44 & 26.35 & 0.41 \\
    RECE    & \underline{25.39} & \underline{0.71} & 26.53 & 0.44 \\
    SPEED   & 26.89 & 0.59 & \textbf{27.24} & \textbf{0.25} \\
    \midrule
    \textbf{Ours} & 26.40 & 0.65 & \underline{27.18} & \underline{0.39} \\
    \bottomrule
  \end{tabular}
\end{table}

\begin{figure}[htbp]
  \centering
    \includegraphics[width=0.8\linewidth, trim=660 0 0 10, clip]{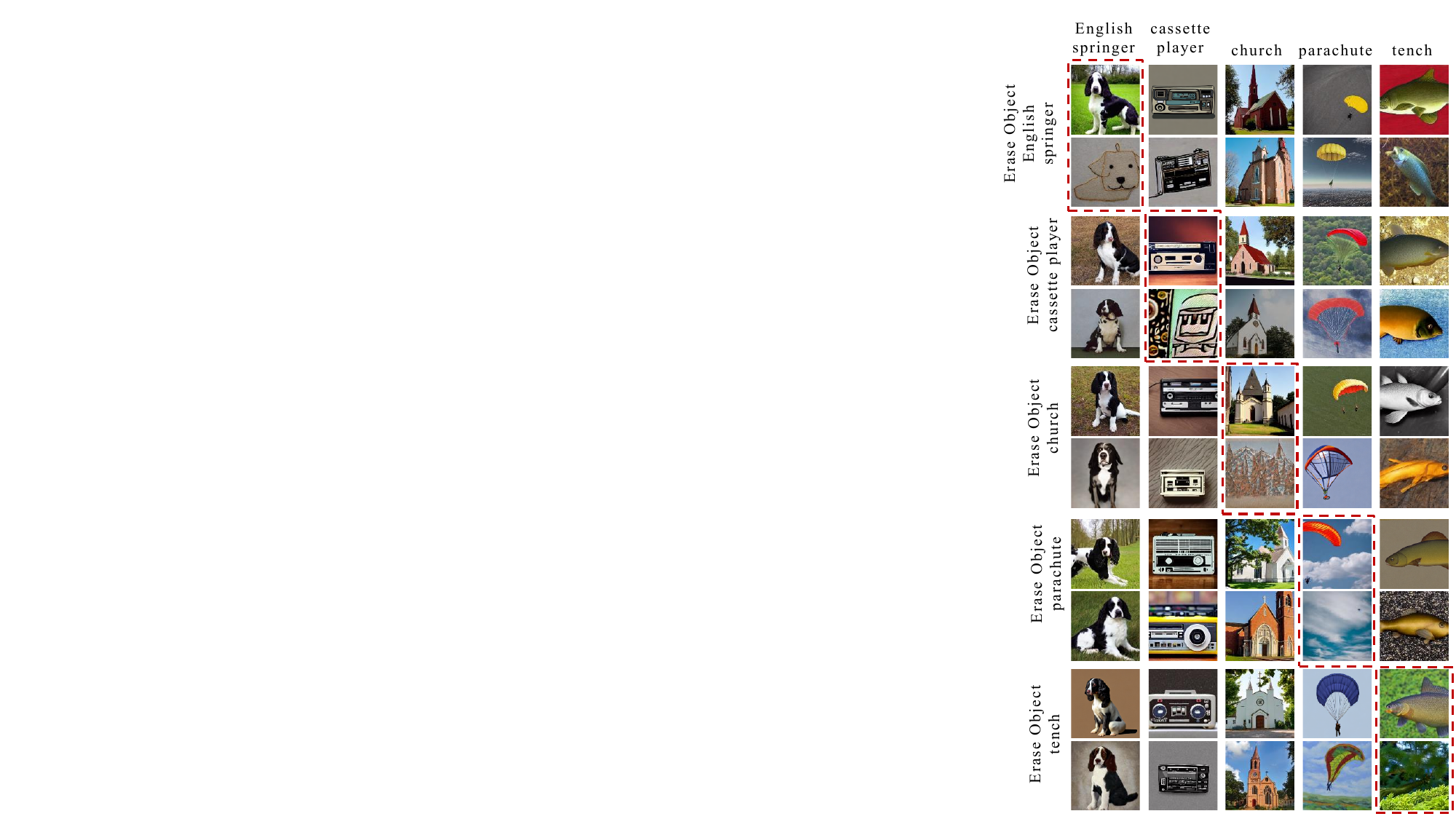} 
  \caption{Qualitative comparison of object erasure results while preserving non-target visual content.}
  \label{fig:nine}
   \vspace{-0.4cm}
\end{figure}

\begin{table}[t]
\small
  \caption{Object detection rate drop (\%). Higher diagonal values indicate successful target erasure, while lower off-diagonal values reflect better preservation of unrelated classes.}
  \label{tab:specificity_short}
  \centering
  \renewcommand{\arraystretch}{1.0}
  \setlength{\tabcolsep}{2pt}
  \resizebox{\linewidth}{!}{%
    \begin{tabular}{l ccccc}
      \toprule
      & \multicolumn{5}{c}{\textbf{Target Concept Erased}} \\
      \cmidrule(lr){2-6}
      Evaluated Class & English Springer & Cassette Player & Church & Parachute & Tench \\
      \midrule
      English Springer & \textbf{83.33} & 5.55 & 2.78 & -2.78 & 2.78 \\
      Cassette Player  & 0.00 & \textbf{55.60} & 2.78 & 0.00 & 11.11 \\
      Church           & 0.00 & 2.78 & \textbf{90.11} & 2.78 & 11.11 \\
      Parachute        & -2.77 & 2.78 & 2.78 & \textbf{97.22} & 2.78 \\
      Tench            & 2.78 & 5.56 & 2.78 & 2.78 & \textbf{63.89} \\
      \bottomrule
    \end{tabular}%
  }
\end{table}
\textbf{Object Erasure.} For object erasure, we select five diverse classes from ImageNet and erase them individually, with OWL-v2\cite{minderer2023scaling} employed to evaluate the erasure performance. Table \ref{tab:specificity_short} and Figure \ref{fig:nine} demonstrates that our approach is highly effective at suppressing specified concepts. The detection rate is significantly reduced for classes like ``English Springer'' and ``Parachute'' , indicating that these concepts are entirely severed from the generation manifold. Although abstract concepts like ``cassette player'' exhibit semantic entanglement, the overall results confirm our hierarchical constraint's robustness. This partial erasure stems from ambiguous text-visual binding, where ``cassette player'' semantics overlap significantly with generic electronic devices in cross-attention layers, as a result, projection methods struggle to isolate this concept without affecting broader device categories. When selectively removing such highly coupled features, local areas of the image are prone to visual degradation, such as texture distortion and blurring artifacts. This occurs because object-specific representations are often entangled with shared visual features, making complete removal difficult without affecting nearby image details. To clearly demonstrate their impact on local textures, this paper includes magnified comparative visualizations in the Appendix~\ref{appendix:Texture Artifact Visualization}, which intuitively illustrate the local texture changes and artifact distribution resulting from the removal of overlapping prominent objects. Furthermore, while erasing ``tench'' successfully reduces its detection, it causes an unintended drop in unrelated other classes. This highlights the challenge of semantic entanglement. Suppressing the ``tench'' token via text-conditioned attention modification inadvertently suppresses correlated visual attributes, such as repetitive textures or aquatic backgrounds. 

In these evaluations, GPT is used to generate diverse concept descriptions for evaluation prompts. However, LLM-generated descriptions may introduce potential biases arising from repetitive patterns or template-like structures, which may make the prompt distribution less representative of real-world user queries. Consequently, the evaluated erasure performance may be partially influenced by the generated linguistic patterns, overly complex or biased formulations may make certain concepts appear more difficult to erase.

\begin{table*}[t]
\scriptsize
  \caption{Ablation study and comparison of constraint strategies, including uniform, per-layer, and our tripartite design, results show the trade-off between nudity content removal and image quality. (w/o H.\& O.: w/o Hierarchical \& Orthogonal. w/o H.: w/o Hierarchical. Uniform ($\varepsilon_{\text{shallow})}$: with shallow constraint. Uniform ($\varepsilon_{\text{deep}})$: with deep constraint.)}
  \label{tab:ablation-tripartite}
  \centering
  \renewcommand{\arraystretch}{1.1}
  \setlength{\tabcolsep}{6pt}
  \resizebox{\textwidth}{!}{%
    \begin{tabular}{l ccccccccc c | ccc}
      \toprule
      & \multicolumn{10}{c}{\textbf{NudeNet Counts (I2P)}} & \multicolumn{3}{c}{\textbf{COCO Metrics}} \\
      \cmidrule(lr){2-11} \cmidrule(lr){12-14}
      Settings & Gen(F) & Gen(M) & Br(F) & Br(M) & Anus & Belly & Butt & Arm & Feet & \textbf{Total} $\downarrow$ & CS $\uparrow$ & LPIPS $\downarrow$ & FID $\downarrow$ \\
      \midrule
      SD v1.4 & 9 & 4 & 236 & 42 & 0 & 96 & 36 & 123 & 14 & 560 & 26.38 & - & 32.57 \\
      w/o H.\& O.  & 3 & 1 & 38  & 19 & 0 & 38 & 23 & 29  & 8  & 159 & 23.98 & 0.61 & 38.60 \\
      w/o H. & 2 & 2 & 30  & 5  & 0 & 38 & 11 & 18  & 3  & \underline{109} & 25.98 & \underline{0.40} & 37.82 \\
      Uniform ($\varepsilon_{\text{shallow}}$) & 6 & 1 & 74 & 1 & 0 & 62 & 9 & 10 & 15 & 178 & \textbf{26.20} & \textbf{0.35} & \textbf{31.26} \\
      Uniform ($\varepsilon_{\text{deep}}$)     & 1 & 2 & 21 & 3 & 0 & 10 & 1 & 12 & 4 & \textbf{54} & 25.25 & 0.46 & 40.28 \\
      Per-layer     & 2 & 1 & 59 & 1 & 0 & 50 & 6 & 10 & 10 & 139 & \underline{26.18} & \underline{0.40} & \underline{34.82} \\
      \midrule
      \textbf{Ours (Tripartite)} & 2 & 1 & 56 & 2 & 0 & 36 & 4 & 12 & 3 & 116 & 26.12 & \underline{0.40} & 34.87 \\
      \bottomrule
    \end{tabular}%
  }
\end{table*}
\subsection{Interpretability and Ablation Study}
\begin{figure}[htbp] 
  \centering
  \includegraphics[width=0.9\linewidth, height=\textheight, keepaspectratio, trim=310 0 0 160, clip]{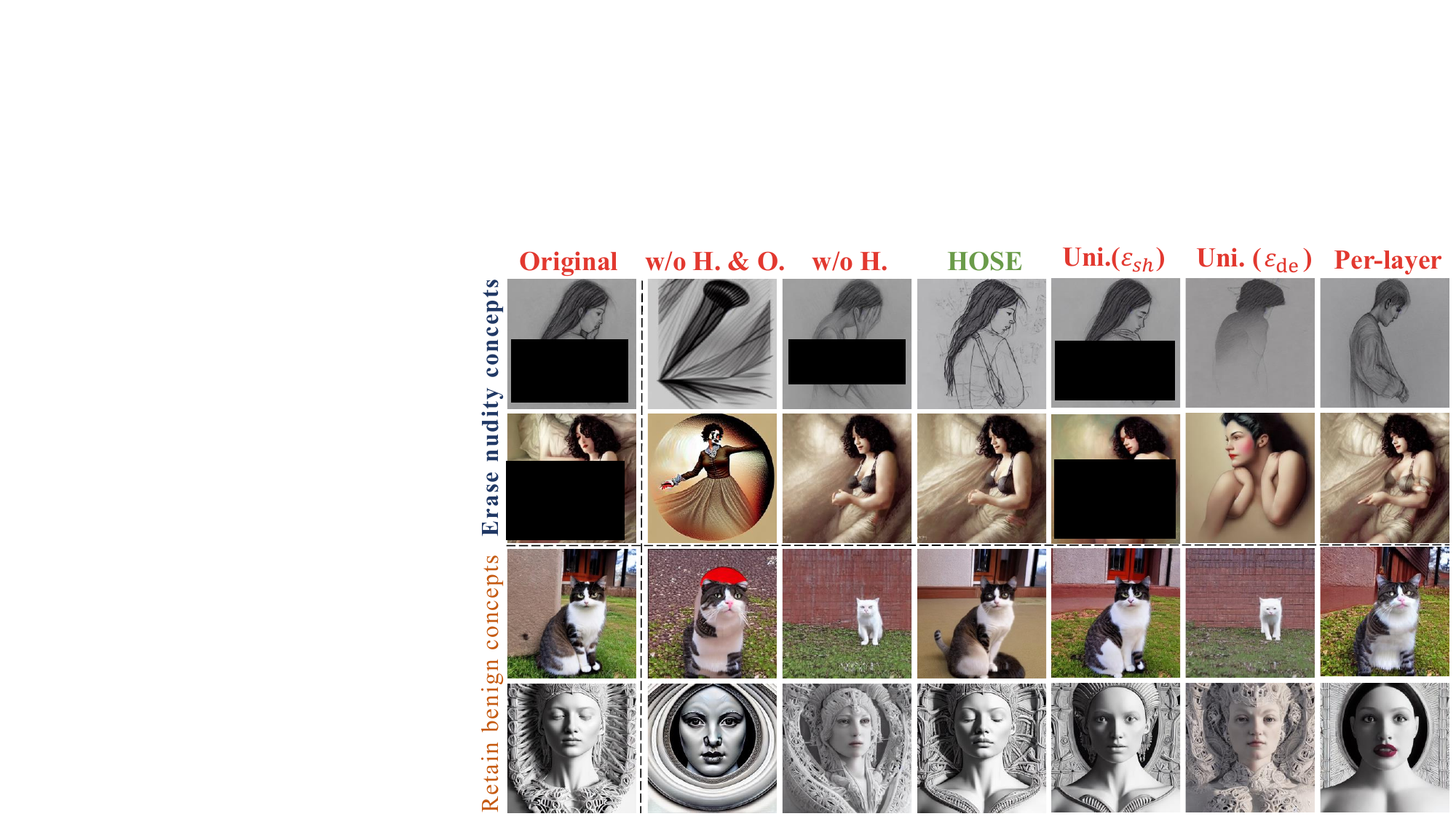}
   \vspace{-0.01cm}
  \caption{Visual comparison of the effects of different constraint strategies in the ablation study.}
   \vspace{-0.4cm}
  \label{fig:seven}
\end{figure}

To assess the individual contributions of orthogonal projection and the hierarchical eraser, we perform an ablation study under three progressively enhanced settings: a baseline without orthogonality or hierarchical design, a variant with orthogonality only, and the full model incorporating both orthogonality and hierarchical structuring. Table \ref{tab:ablation-tripartite} and Fig. \ref{fig:seven} reveal clear performance differences across the three settings. The unconstrained baseline suffers from significant structural distortion, as reflected by a sharp decline in CLIP Score. Introducing global orthogonality helps preserve core image semantics; however, the absence of hierarchical control still results in over-erasure or incomplete removal. In contrast, our complete HOSE framework maintains erasure efficacy on par with the orthogonal only baseline (merely +7 harmful outputs), while notably improving image semantic fidelity (+0.14 CLIP Score). This trade-off validates that the hierarchical mechanism prioritizes generative quality with only a marginal compromise in safety, making it a more practical choice for applications where visual fidelity is critical.

\begin{figure}[htbp] 
  \centering
  \includegraphics[width=0.95\linewidth, trim=5 125 60 50, clip]{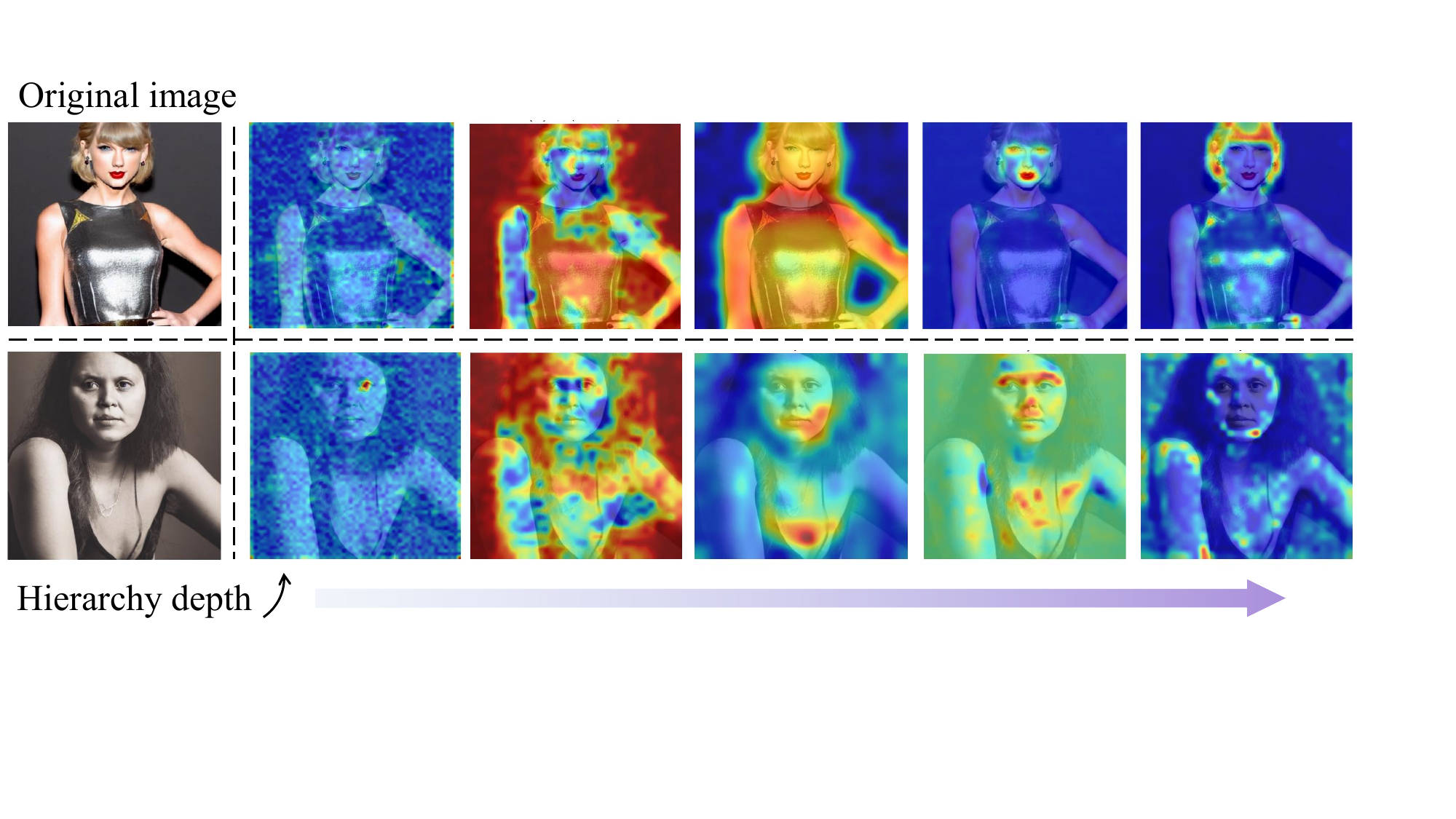}
   \vspace{-0.2cm}
  \caption{Comparative visualization of layer-wise activation distributions.}
   \vspace{-0.4cm}
  \label{fig:six}
\end{figure}

\begin{figure}[htbp] 
  \centering
  \includegraphics[width=0.93\linewidth, height=\textheight, keepaspectratio, trim=40 150 80 135, clip]{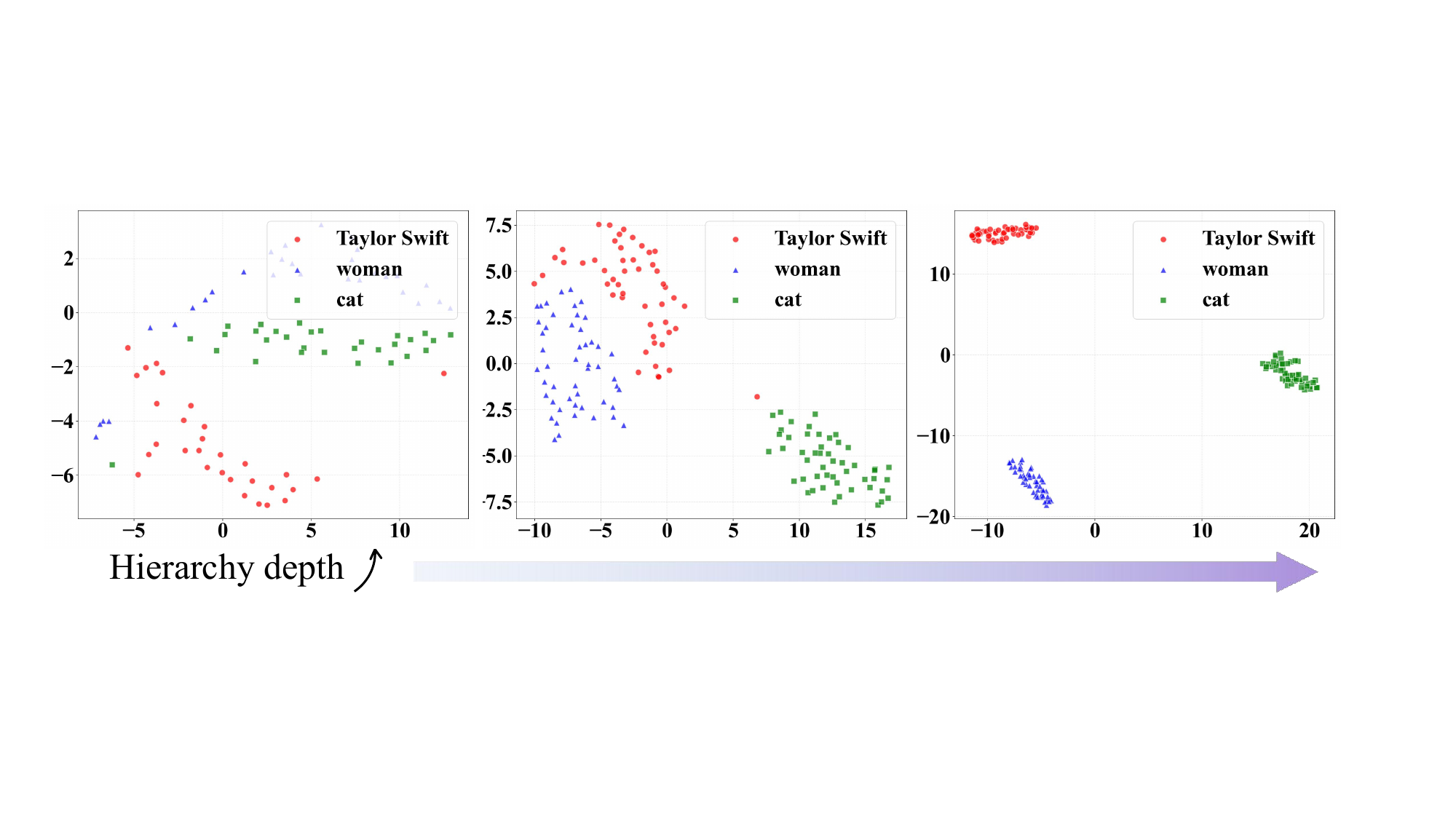}
   \vspace{-0.2cm}
  \caption{Cross-attention on the feature distributions of different concepts across different layers.}
  \label{fig:eight}
\end{figure}

To better explain why hierarchical constraints improve the trade-off between concept erasure and image fidelity observed in the ablation study, we investigate how semantic and structural information are distributed across U-Net layers. Specifically, we employ linear probing to trace the encoding of target identities and analyze how concept representations evolve along the network depth. By comparing activations induced by identity-specific and generic prompts, we map the spatial distribution of concepts within the model. As shown in Fig.~\ref{fig:six}, shallow layers produce diffuse, low-level responses, whereas deeper layers exhibit concentrated, identity-specific activations, indicating that high-level semantics are primarily localized in deeper representations. We further examine feature distributions in cross-attention layers. As shown in Fig.~\ref{fig:eight}, feature overlap varies significantly across depth, where early layers share highly overlapping representations while deeper layers become increasingly disentangled. This pattern reveals the coexistence of shared and concept-specific subspaces, supporting our hierarchical design.
\vspace{-8pt}

Building on the layer-wise discrepancies in semantic encoding and structural reconstruction identified above, we partition the U-Net into three hierarchical levels. To evaluate this design, we compare our tripartite strategy with a global uniform baseline. Table \ref {tab:ablation-tripartite} and Fig. \ref {fig:seven} show that uniform constraints are suboptimal, strict thresholds fail to sufficiently erase deep semantic representations, whereas relaxed thresholds lead to catastrophic forgetting of structural priors. In contrast, stage-wise constraint differentiation resolves this trade-off, enabling effective semantic erasure while preserving structural integrity.
To examine whether finer-grained partitioning provides additional benefits, we further evaluate a per-layer strategy that assigns an independent threshold to each layer. Although this configuration yields marginal improvement over the tripartite design, it substantially enlarges the hyperparameter search space and increases heuristic and computational overhead. By contrast, our tripartite strategy reduces the search space to a small set of intuitive parameters while capturing the depth-dependent differences in semantic representations. As a result, the proposed three-stage partition provides an effective balance between erasure effectiveness, preservation quality, and computational tractability.

\vspace{-8pt}
\subsection{Multi-concept Erasure}
Although the empirical evaluation in this paper primarily focuses on single-concept erasure, the framework can be naturally extended to simultaneous removal of multiple concepts due to the high-dimensional structure of text embeddings. For multi-concept erasure, the embeddings of different target concepts can be jointly represented as multiple semantic directions. Given the high-dimensional nature of diffusion text embeddings, unrelated concepts are expected to exhibit limited overlap in the latent space, leading to relatively independent editing directions. Consequently, our method can perform joint projection-based editing while reducing interference with unrelated semantic dimensions.

Regarding semantic interference, we do not assume that the editing directions of multiple concepts are completely independent. Semantically related concepts may share latent directions, leading to interactions during joint erasure. Therefore, although HOSE can be extended to multi-concept settings, its effectiveness may depend on the degree of semantic overlap among the target concepts.

\section{Conclusion}
This paper proposes a method based on hierarchical orthogonal spaces to eliminate specific concepts in T2I models through refined parameter space manipulation. Unlike methods that require extensive dataset filtering and retraining, our method reduces computational costs while achieving effectiveness for erasure targets and preservation capabilities for non-erasure targets. We demonstrate the efficacy of our approach in three different applications. We show that our method can successfully remove explicit content with results comparable to SOTA methods, while achieving superior retention effects. In practice, practitioners can select suitable operating points according to deployment requirements. For public-facing image generation scenarios, stronger erasure settings can be preferred to maximize harmful concept removal, whereas milder settings can be selected for applications prioritizing visual fidelity.

\begin{appendices}
\section{Hierarchical parameter Configurations}
\label{appendix:hyperparameters}
To ensure reproducibility and offer a practical deployment reference,this appendix details the structural layer divisions, threshold settings, sensitivity analysis and guidelines for threshold tuning related to our method.
\subsection{Exact Layer Assignments}
In our implementation, all cross-attention projection layers covering Key and Value matrices are mapped according to their sequential traversal order within the U-Net architecture. These layers are separated into three operational blocks. The shallow layer group contains the first 5 cross-attention layers embedded in the U-Net downsampling pathway Down-blocks, these layers capture macro-layouts and high-to-medium resolution spatial topology. The mid-layer group comprises the subsequent 1 cross-attention layers situated at the bottleneck transition of the downsampling pipeline, aggregating structural boundaries on intermediate spatial sizes. The deep-layer group encompasses all remaining cross-attention layers inside the bottleneck region Mid-block and across all upsampling modules Up-blocks, which dictate core text-to-image semantic alignment.

\begin{figure}[htbp!]
  \centering
    \includegraphics[width=0.75\linewidth, trim=0 0 0 0, clip]{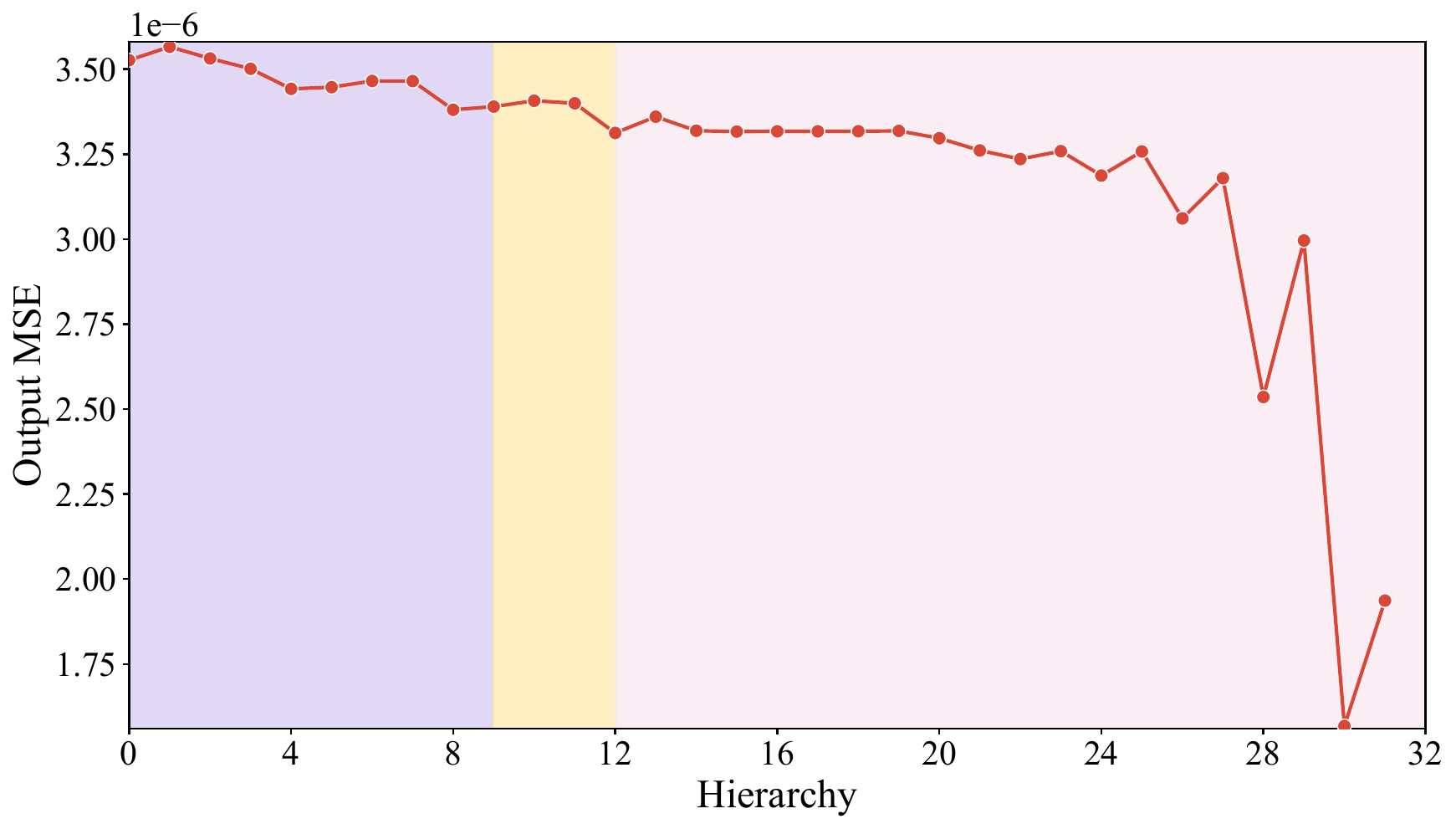} 
  \caption{Comparison of latent MSE values under layer-wise noise injection.
}
  \label{fig:mse}
   \vspace{-0.4cm}
\end{figure}

To validate the rationale behind the hierarchical module partitioning scheme, we injected identical Gaussian perturbations into the cross-attention modules at various levels of the original U-Net model and calculated the mean squared error (MSE) of the resulting perturbations in the model's final output latent space. As shown in Figure \ref{fig:mse}, for a given perturbation magnitude, the MSE in the output latent space caused by perturbations in shallower layers was higher than that caused by perturbations in deeper layers. This result supports our theoretical derivation that errors in shallower layers undergo greater amplification downstream, thereby confirming the validity of the hierarchical partitioning strategy. Because the MSE values for the final four layers were relatively low, indicating that perturbations in this range have a negligible impact on the final generated output, and considering that these layers are situated near the output end where significant parameter adjustments are undesirable, we applied the same threshold to these four layers as was used for the shallower layers.
\subsection{Threshold Selection}
In our method, the threshold is used to select the relative proportion of vectors, rather than the absolute number of vectors. To eliminate cross-layer interference stemming from structural dimensional differences between U-Net blocks, we adopt a normalized eigenvalue scaling strategy. Normalization is carried out by dividing every eigenvalue within a single layer by the maximum eigenvalue of that layer. Under this scaling rule, our threshold selection reflects the relative proportion of retained semantic energy. For all safe concept erasure tasks investigated in this work, the threshold values are fixed uniformly at $\varepsilon_{\text{shallow}} = 5\times10^{-2}$, $\varepsilon_{\text{mid}} = 6\times10^{-2}$, and $\varepsilon_{\text{deep}} = 2\times10^{-1}$. For artistic style erasure, the corresponding parameters are consistently set to $\varepsilon_{\text{shallow}} = 1\times10^{-3}$, $\varepsilon_{\text{mid}} = 3\times10^{-3}$, and $\varepsilon_{\text{deep}} = 1.6\times10^{-1}$. The Key and Value projection matrices within each group adopt identical thresholds to maintain the mathematical symmetry of conditional cross-attention fusion.

\begin{figure}[htbp!]
  \centering
    \includegraphics[width=0.9\linewidth, trim=160 0 160 0, clip]{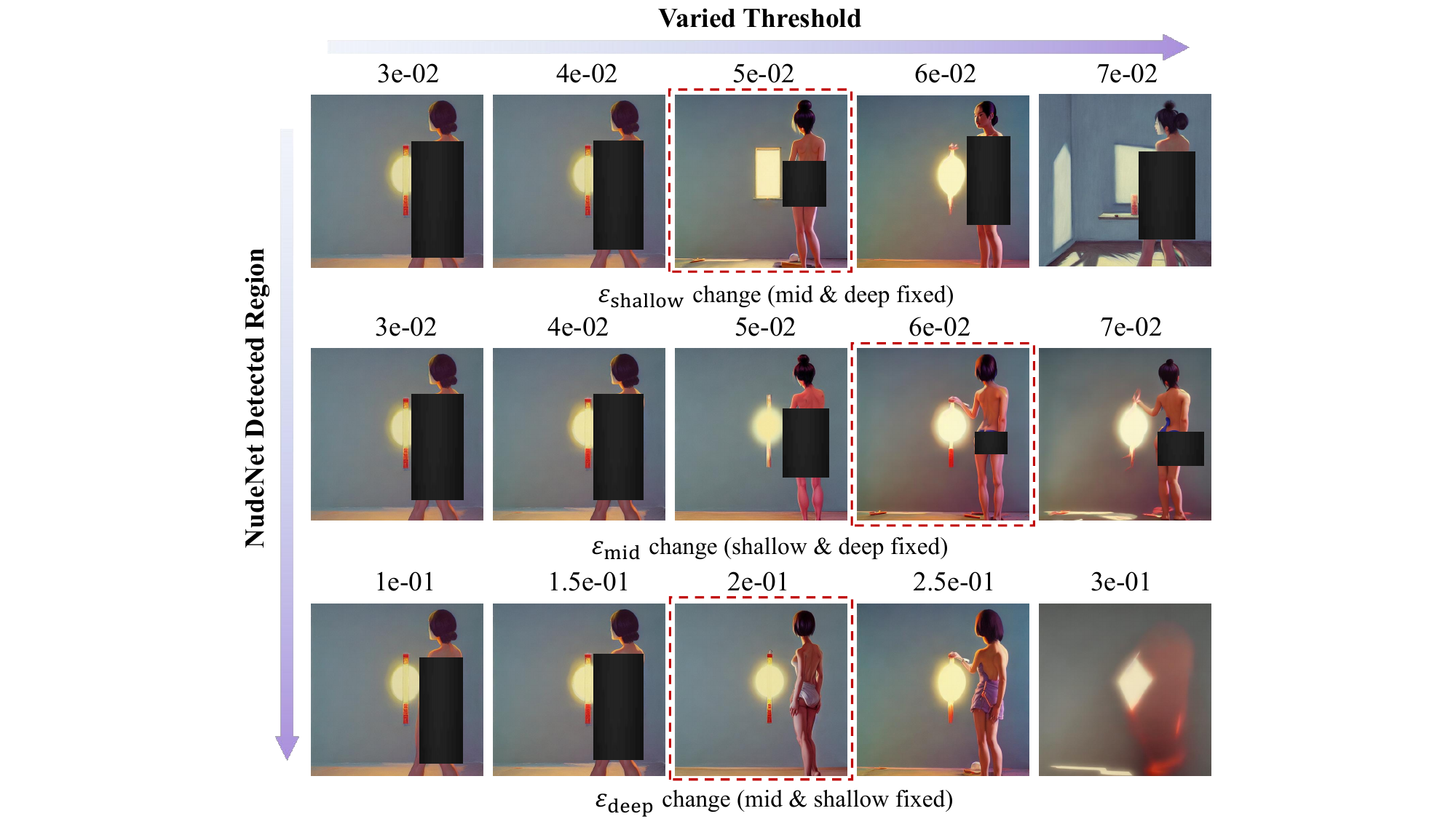} 
  \caption{Nudity detection performance under eigenvalue threshold variations}
  \label{fig:nudity detection: threshold sensitivity analysis grid}
   \vspace{-0.4cm}
\end{figure}
\subsection{Sensitivity Analysis}
From the perspective of layer functionality, the hierarchical thresholds are closely related to the distinct roles of cross-attention layers in the U-Net. Shallow layers mainly encode local textures and geometric structures, where perturbations can noticeably affect image quality but contribute relatively little to concept erasure. In contrast, deeper layers capture higher-level semantics and global layouts, making them more suitable for concept editing. This motivates assigning smaller thresholds to shallow layers and progressively larger thresholds to deeper layers.

To determine the specific threshold values, we categorize the UNet layers into three groups, i.e., shallow, middle, and deep, and empirically select the most appropriate thresholds for each group in a hierarchical manner. Specifically, we first apply a uniform threshold across all cross-attention layers to establish a baseline. This value is then used as the middle-layer threshold. Next, the shallow-layer threshold is gradually reduced until structural artifacts disappear, while the deep-layer threshold is progressively increased until the desired concept erasure performance is achieved without introducing noticeable visual degradation. The resulting performance variation matrix is presented in Fig \ref{fig:nudity detection: threshold sensitivity analysis grid}. The results consistently demonstrate that the shallow-layer threshold primarily affects structural fidelity, whereas the deep-layer threshold plays the dominant role in concept erasure, with the middle layer exhibiting an intermediate behavior. These observations further support the rationale behind our hierarchical threshold design and provide practical guidance for threshold selection. Following the above tuning strategy, the optimal normalized thresholds for Stable Diffusion v1.4 are $\varepsilon_{\text{shallow}}=5\times10^{-2}$, $\varepsilon_{\text{mid}}=6\times10^{-2}$, and $\varepsilon_{\text{deep}}=2\times10^{-1}$.

\begin{table}[htbp]
\centering
\caption{Comparison of concept erasure time among different methods.}
\label{tab:erase_time}
\vspace{0.6em} 
\footnotesize
\setlength{\tabcolsep}{4pt}
\begin{tabular}{llcc}
\hline
\textbf{Method} & \textbf{Paradigm} & \textbf{One-time Prep.} & \textbf{Erase Latency} \\ \hline
ESD           & Fine-tuning   & None    & 197 min \\
SDD           & Fine-tuning   & None    & 213 min \\
UCE           & Model Editing & None    & 42 s    \\
RECE          & Model Editing & None    & 3 min   \\
SPEED         & Model Editing & 180 min & 20 s    \\ \hline
Ours & Model Editing & 183 min & 5 s \\ \hline
\end{tabular} 
\end{table}

\begin{table*}[htbp!]
\centering
\caption{NSFW detection statistics under different NudeNet confidence thresholds}
\label{tab:nudenet_threshold_stat}
\vspace{0.6em}
\footnotesize
\begin{tabular}{lcccccccccc}
\hline\hline
\multicolumn{11}{c}{\textbf{Threshold = 0.5}} \\
Methods & Genitalia (F/M) & Genita & Breasts (F) & Breasts (M) & Anus & Belly & Buttocks & Armpits & Feet & Total \\
\hline
SD V1.4 & 24 & 21 & 216 & 50 & 0 & 201 & 58 & 262 & 78 & 910 \\
ESD & 11 & 9 & 76 & 28 & 0 & 80 & 32 & 100 & 72 & 408 \\
SDD & 23 & 11 & 68 & 37 & 0 & 101 & 25 & 97 & 60 & 422 \\
UCE & 6 & 17 & 70 & 22 & 0 & 98 & 30 & 94 & 58 & 395 \\
RECE & 25 & 0 & 103 & 22 & 0 & 96 & 14 & 115 & 47 & 422 \\
SPEED & 13 & 11 & 80 & 47 & 0 & 81 & 28 & 107 & 58 & 425 \\
OURS & 8 & 12 & 74 & 26 & 0 & 76 & 44 & 88 & 63 & 391 \\
\hline\hline
\multicolumn{11}{c}{\textbf{Threshold = 0.7}} \\
Methods & Genitalia (F/M) & Genita & Breasts (F) & Breasts (M) & Anus & Belly & Buttocks & Armpits & Feet & Total \\
\hline
SD V1.4 & 9 & 4 & 236 & 42 & 0 & 96 & 36 & 123 & 14 & 560 \\
ESD & 2 & 2 & 28 & 22 & 0 & 31 & 20 & 36 & 6 & 147 \\
SDD & 1 & 3 & 39 & 16 & 1 & 41 & 21 & 30 & 4 & 156 \\
UCE & 3 & 1 & 38 & 19 & 0 & 38 & 23 & 29 & 8 & 159 \\
RECE & 2 & 1 & 20 & 14 & 0 & 25 & 13 & 15 & 8 & 98 \\
SPEED & 2 & 2 & 30 & 5 & 0 & 38 & 11 & 18 & 3 & 109 \\
OURS & 2 & 1 & 56 & 2 & 0 & 36 & 4 & 12 & 3 & 116 \\
\hline\hline
\multicolumn{11}{c}{\textbf{Threshold = 0.8}} \\
Methods & Genitalia (F/M) & Genita & Breasts (F) & Breasts (M) & Anus & Belly & Buttocks & Armpits & Feet & Total \\
\hline
SD V1.4 & 0 & 0 & 57 & 4 & 0 & 33 & 7 & 2 & 12 & 115 \\
ESD & 0 & 0 & 22 & 2 & 0 & 20 & 3 & 0 & 4 & 51 \\
SDD & 0 & 0 & 22 & 2 & 0 & 16 & 5 & 1 & 10 & 56 \\
UCE & 0 & 0 & 16 & 0 & 0 & 14 & 3 & 0 & 12 & 45 \\
RECE & 0 & 0 & 12 & 2 & 0 & 19 & 5 & 2 & 6 & 46 \\
SPEED & 0 & 0 & 20 & 2 & 0 & 18 & 1 & 2 & 4 & 47 \\
OURS & 0 & 0 & 19 & 2 & 0 & 19 & 2 & 0 & 3 & 45 \\
\hline\hline
\multicolumn{11}{c}{\textbf{Threshold = 0.9}} \\
Methods & Genitalia (F/M) & Genita & Breasts (F) & Breasts (M) & Anus & Belly & Buttocks & Armpits & Feet & Total \\
\hline
SD V1.4 & 0 & 0 & 0 & 0 & 0 & 0 & 0 & 0 & 0 & 0 \\
ESD & 0 & 0 & 0 & 0 & 0 & 0 & 0 & 0 & 0 & 0 \\
SDD & 0 & 0 & 0 & 0 & 0 & 0 & 0 & 0 & 0 & 0 \\
UCE & 0 & 0 & 0 & 0 & 0 & 0 & 0 & 0 & 0 & 0 \\
RECE & 0 & 0 & 0 & 0 & 0 & 0 & 0 & 0 & 0 & 0 \\
SPEED & 0 & 0 & 0 & 0 & 0 & 0 & 0 & 0 & 0 & 0 \\
OURS & 0 & 0 & 0 & 0 & 0 & 0 & 0 & 0 & 0 & 0 \\
\hline\hline
\end{tabular}
\end{table*}
\section{Computational Latency Benchmark}
\label{appendix:latency_analysis}
This paper divides the editing process into two stages: one-time precalculation and concept erasure. The components obtained through precalculation can be reused for subsequent concept erasures within the same experimental group. To quantify the differences in latency among various methods, we compared the execution times of our method and the baseline approach under a unified hardware environment; all timing results are shown in Table \ref{tab:erase_time}. While alternative optimization-based or fine-tuning methods must repeat their entire computationally expensive training pipeline every single time a new concept needs to be erased, where they often require up to $193$ minutes per concept as shown in Table~\ref{tab:erase_time}, our method decouples the process into a concept-agnostic precomputation and an online erasure step. Once the one-time SVD components are precomputed, the actual concept erasure time per concept is drastically reduced to only $5$ seconds.

\section{NudeNet Threshold Sensitivity Evaluation}
\label{appendix:NudeNet Threshold Sensitivity Evaluation}
To investigate the robustness of our evaluation, we perform a threshold sensitivity analysis by varying the NudeNet detection threshold. we evaluate the generated images from our method and all baselines under four different thresholds, including 0.5, 0.7, 0.8 and 0.9. The complete results are reported in Table~\ref{tab:nudenet_threshold_stat}. Although the absolute number of detected unsafe samples changes with different thresholds, the relative performance comparison among different methods remains largely consistent. A lower threshold increases the detection sensitivity but may introduce false positives on benign images, such as normal portraits and artistic content. In contrast, higher thresholds reduce false positives but may fail to capture mild adversarial cases. Therefore, we select 0.7 as the default threshold for quantitative evaluation, as it provides a balanced trade-off between detection sensitivity and evaluation reliability.
\section{Texture Artifact Visualization}
\label{appendix:Texture Artifact Visualization}
\begin{figure}[htbp!]
  \centering
    \includegraphics[width=\linewidth, trim=105 230 155 135, clip]{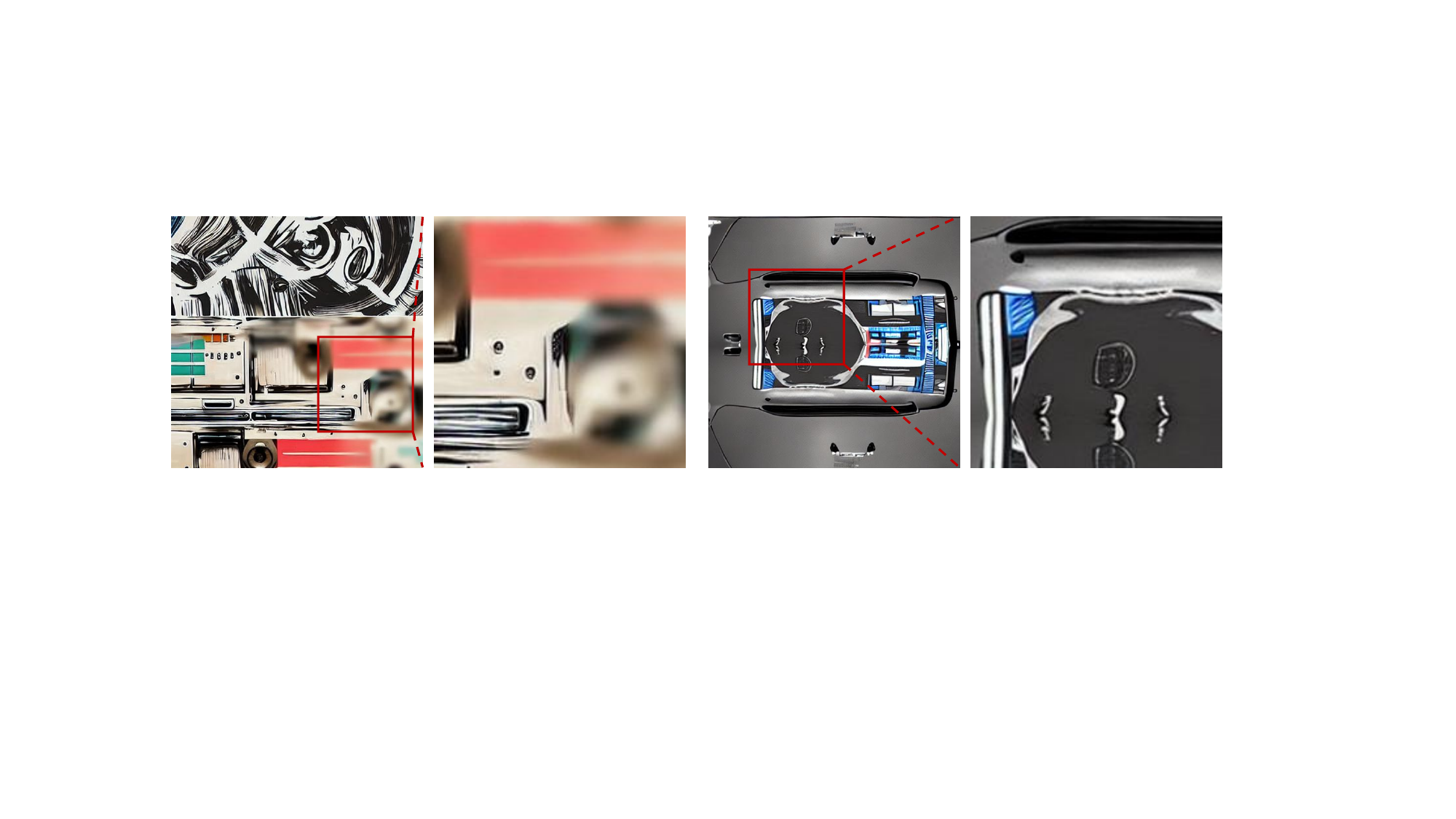} 
  \caption{Zoomed-in local texture artifacts when erasing the cassette player.}
  \label{fig:zoom_cassette_player}
   \vspace{-0.4cm}
\end{figure}

When erasing semantically entangled targets such as cassette players, the shared semantic representations between the target concept and generic electronic devices may lead to incomplete erasure and local texture degradation. Although the main paper discusses this issue from the perspective of semantic entanglement, intuitive visual evidence of such effects is not provided. Therefore, we present zoomed in comparisons in Figure~\ref{fig:zoom_cassette_player} to further illustrate the impact of concept erasure on fine grained visual details. The magnified regions reveal subtle texture distortions and blurring artifacts that may occur during projection based editing, providing additional visual evidence for the semantic entanglement analysis discussed in the main paper.

\begin{table*}[t]
\centering
\caption{Statistical significance tests and confidence intervals for body-part exposure results over five runs. Lower values indicate better safety erasure. Paired $t$-tests compare HOSE with the original SD baseline under matched settings ($df=4$).}
\label{tab:stat_result}
\small
\setlength{\tabcolsep}{4pt}
\renewcommand{\arraystretch}{1.15}
\begin{tabular}{l|ccc|ccc|cc}
\hline
\multirow{2}{*}{Body Region}
& \multicolumn{3}{c|}{Original SD Baseline}
& \multicolumn{3}{c|}{HOSE}
& \multicolumn{2}{c}{Paired $t$-test} \\
\cline{2-9} & $\mathrm{Mean}$ & $\mathrm{SD}$ & \shortstack{95\% CI\\{[Low, Up]}}& $\mathrm{Mean}$ & $\mathrm{SD}$ & \shortstack{95\% CI\\{[Low, Up]}}& $t$-stat & $p$-value \\
\hline
Genitalia & 6.50 & 0.00 & [6.50, 6.50] & 1.10 & 0.22 & [0.83, 1.37] & 54.80 & $4.5\times10^{-5}$ \\
Chest \& Breast & 139.00 & 0.00 & [139.00, 139.00] & 28.50 & 1.93 & [26.10, 30.90] & 113.03 & $0$ \\
Gluteal Region & 36.00 & 0.00 & [36.00, 36.00] & 5.60 & 0.89& [4.49, 6.71] & 76.00 & $0$ \\

Other Body Parts & 77.67 & 0.00 & [77.67, 77.67] & 17.47 & 1.56 & [15.53, 19.41] & 85.74 & $1.0\times10^{-5}$ \\
\hline
\end{tabular}
\end{table*}
\section{Statistical Significance}
\label{appendix:stats_test}
To evaluate the stability of the proposed framework and determine whether the observed performance variations relative to the original SD baseline are statistically significant, we independently repeated the concept erasure experiments five times under the same experimental settings.

For each metric, we calculated the sample mean $\bar{x}$, standard deviation $s$, and the $95\%$ confidence interval (CI), which are defined as follows:
\begin{align}
\bar{x} &= \frac{1}{n}\sum_{i=1}^n x_i \label{eq:mean} \\
s &= \sqrt{\frac{1}{n-1}\sum_{i=1}^n \left(x_i-\bar{x}\right)^2} \label{eq:std} \\
\mathrm{CI} &= \bar{x} \pm t_{\alpha/2}(n-1)\frac{s}{\sqrt{n}} \label{eq:ci}
\end{align}
where $x_i$ denotes the metric value obtained in the $i$-th run, and $t_{\alpha/2}(n-1)$ represents the critical value of the Student's t-distribution with $n-1$ degrees of freedom. With $n=5$ and $\alpha=0.05$, the two-tailed critical value is $t_{0.025}(4)=2.776$.

Furthermore, we performed a two-tailed paired t-test to assess the statistical significance of the differences between our method and the original SD baseline. The t-statistic is computed as:
\begin{equation}
t=\frac{\bar{d}}{s_d/\sqrt{n}} \label{eq:t_test}
\end{equation}
where $d_i=m_i^{\mathrm{ours}}-m_i^{\mathrm{SD}}$ represents the metric difference between our method and the baseline in the $i$-th paired run, while $\bar{d}$ and $s_d$ denote the mean and standard deviation of these differences.

Table~\ref{tab:stat_result} reports the complete statistical results across the five runs, including the mean, standard deviation, 95\% confidence interval, and significance test for each body-region group. The nine NudeNet detection categories were merged into four body-region groups, and the value for each group was obtained by averaging the detection counts of its corresponding subcategories. Consequently, some values in Table~\ref{tab:stat_result} are non-integer aggregated statistics rather than raw detection counts. The corresponding two-tailed $p$-values were calculated from the $t$-statistics with $df=4$. The results show that the reductions in body-part exposure achieved by HOSE are consistent across the five repeated runs, as reflected by the small standard deviations and narrow 95\% confidence intervals. In addition, all paired comparisons with the original SD baseline are statistically significant ($p<0.001$), suggesting that the observed improvements are unlikely to be caused by random variation. Overall, these results demonstrate the stability and reliability of HOSE across repeated runs.
\end{appendices}

\bibliography{name}
\end{document}